# Number game –Experience of a European research infrastructure (CLARIN) for the analysis of web traffic

**Go Sugimoto**
ACDH-OEAW
Vienna, Austria
go.sugimoto@oeaw.ac.at

## Abstract

CLARIN (Common Language Resources and Technology Infrastructure) is regarded as one of the most important European research infrastructures, offering and promoting a wide array of useful services for (digital) research in linguistics and humanities. However, the assessment of the users for its core technical development has been highly limited, therefore, it is unclear if the community is thoroughly aware of the status-quo of the growing infrastructure. In addition, CLARIN does not seem to be fully materialised marketing and business plans and strategies despite its strong technical assets. This article analyses the web traffic of the Virtual Language Observatory, one of the main web applications of CLARIN and a symbol of pan-European research cooperation, to evaluate the users and performance of the service in a transparent and scientific way. It is envisaged that the paper can raise awareness of the pressing issues on objective and transparent operation of the infrastructure though *Open Evaluation*, and the synergy between marketing and technical development. It also investigates the "science of web analytics" in an attempt to document the research process for the purpose of reusability and reproducibility, thus to find universal lessons for the use of a web analytics, rather than to merely produce a statistical report of a particular website which loses its value outside its context.

## 1 Background

When creating a website, nobody is free from numbers. Web creators need to know the amount of access to evaluate the popularity of the website, to monitor the server load and security issues, or to assess the revenue streams of online shops. In general, numbers are objective and useful to indicate where you are and where you would like to go.

CLARIN is one of the leading initiatives of scientific research on natural languages in Europe. It started from a European Commission funded project and is now managed under a legal entity, CLARIN-ERIC. This set-up allows them to be recognised as a solid research infrastructure, an ESFRI (the European Strategy Forum on Research Infrastructures) landmark with 18 official national consortia from 17 EU countries. Alongside the well-established social network of experts, the strength of CLARIN lies in its technical properties. It has developed a wide array of web services to promote and enhance e-research for language experts alike. For example, it offers Virtual Language Observatory (VLO)[1] to discover language resources and tools provided by a large number of data providers. Weblicht[2] is a web application equipped with easy-to-use Natural Language Processing tools. With Content Search[3], one can make a query over several corpora by means of federated search. Furthermore, Federated Identity service[4] enables the users to access CLARIN's password-protected resources via Single Sign-On with academic credentials.

Despite the successful deployment of technical research infrastructure, there seems to be a lack of awareness or understanding in terms of the objective assessment of the infrastructure. In particular, available literatures provide very limited information about the users for the development of its services and clarity on the CLARIN's performance. This paper will, therefore, address the heart of those missing links by providing the statistics of the web traffic to display a snippet of the CLARIN's accountability. The value of web analytics in the enhancement of online services is acknowledged by many publications in a similar environment such as digital libraries and archives (Marek 2011; Stuart 2014; Szajewski 2013). This is, therefore, a vital step for CLARIN to confront with, because the question now is whether its community will continue developing its infrastructural services without adequately and systematically assessing its use, or not.

[1] https://vlo.clarin.eu/
[2] http://weblicht.sfs.uni-tuebingen.de/weblichtwiki/index.php/Main_Page
[3] https://www.clarin.eu/content/content-search
[4] https://www.clarin.eu/content/federated-identity

The section 2 describes more detailed motivations of this paper with references to the policies and trends in the context of European research. The section 3 presents the analysis of the VLO web traffic as the main content of this paper. The section 4 needs more explanation, because it is a by-product of this research and will investigate the behind-the-scenes of web analytics. Unsatisfied with a myriad of "blind" statistical reports on web visitors and the following conclusions which cannot be applied to other websites, the paper tries to shed light upon the documentation of "science of web analytics" as a subject of research, sharing and discussing the process, methodology, and experience of analyses and interpretations, which hopefully brings a few lessons for the research infrastructure and Digital Humanities fields at large. For the purpose of research reusability and reproducibility, the characteristics of the (verbose) documentation can be seen throughout the section 3. The last section summarises the whole discussion and end with conclusions.

# 2 Open evaluation –users and transparent European research

## 2.1 User centric approach for marketing and development

The European Commission described in the principles for access to research infrastructures:

> *"Research Infrastructures should have a policy defining how they regulate, grant and support Access to (potential) Users from academia, business, industry and public services"* (European Commission and Directorate-General for Research and Innovation 2016).

Coincided with the principles, it is appreciated that CLARIN has a strong backing from the linguistic community in Europe and has successfully undertaken a series of projects over the last few years, demonstrating the value of academic and research infrastructure in its own right. However, in spite of the emphasis on the users in the principles, the user evaluation in CLARIN has been rather minimal. In other words, the infrastructure development mostly concentrates on planning and doing in the sense of PDCA[5] and checking and acting are highly limited. The most recent publication (Eckart et al. 2015) reports on the user behaviour of VLO, one of the flagship services of CLARIN, but it does not offer any previous literatures on the subject. Although there are some user evaluations, they seem to be limited in the form of internal community feedback (Goosen and Eckart 2014; Haaf et al. 2014). Wynne (2015b) conducted an analysis on various types of users and target domains, but there seem to be several contradictions on the target users and his conclusions are a bit of overstatement without providing proper evidences to prove them. This paper is intended to present objective statistics and make assumptions and conclusions in a more scientific manner.

Customers (in our case, users) are undoubtedly the protagonist of marketing, but it would be agreed that marketing (in the broadest sense including non-commercial one) is not regarded as the strongest component of the CLARIN operation. CLARIN's so-called user involvement and outreach are the nearest fields of this kind. However, again CLARIN focuses too much on the implementation of the planning and doing of PDCA. The members are very active to organise and participate in national and international activities including conferences, lectures, and workshops, as well as to publish papers and promotional materials, demonstrate software, and post comments and photos in social media. In contrast, the essence of PDCA, checking and acting are not sufficient for effective management. Although there are many ideas to measure and evaluate the impact of marketing outputs (Wynne 2015a), the user involvement group of CLARIN merely records the events and products and counts the number of participants or publications[6]. But, the impact of before and after the outreach activities of any kind, preferably demonstrated in the form of objective numbers can be measured and follow-up actions should be taken to improve the outreach activities. Next to (and in connection to) the impact study of the iterative website development, this task is very straightforward with web analytics. For instance, in the field of humanities, previous research includes one from Fang (2007) who investigated the use of digital archival content with Google Analytics, and another from Szajewski (2013) who explored the impact of improvement for library web content and design. The latter also collected a broad range of literatures on web analytics produced in the field of digital libraries and archives. However he also stressed the significant lack of scholarship dedicated to increase the visibility, discovery, and use of digitised archival assets among a broad audience of users. Indeed, he continued to say that people ignored the value of web analytics in informing development of web outreach programmes. In the meanwhile, Hoďáková and Neméthová (2011) highlighted some of the important scopes of web analytics in marketing, for example, presenting the definition, and the relation to marketing mix, as well as segmentation and Return on Investment (RoI) of the activities on the internet, in relation to an in-depth statistical analysis of a school magazine website. It is understandable that CLARIN may not have enough resources to conduct an intensive marketing, but, in this respect too, a web analytics offers a cost-effective option.

[5] https://en.wikipedia.org/wiki/PDCA
[6] https://www.clarin.eu/user-involvement

Marketing is a vehicle to develop business strategies and its methodologies are more and more adopted for non-profit organisations. In fact, recently CLARIN PLUS project published a value proposition, specifying various stakeholders in connection to their interests of CLARIN infrastructural services (Maegaard et al. 2016). It is a valuable document in its own right, but, to reaffirm their statements, the marketing perspective of web analytics could play a promising role for CLARIN to ensure its importance of research infrastructure, as digital/web analytics is an integral part of core business strategies (Digital Analytics Association. no date).

## 2.2 Trend of transparent and measureable research

The ESFRI Roadmap outlines the RoI as follows:

> *"The ESFRI Landmarks need continuous support for successful completion, operation and upgrade in line with the optimal management and maximum return on investment."* (European Strategy Forum on Research Infrastructures 2016).

For big research sponsors such as the European Commission and national governments, spending millions of Euros every year, it is a long standing, as well as a burgeoning question in the era of a new digital paradigm, on how to ensure efficient management and sufficient RoI. In addition, the research project calls of the European Commission nowadays require a great deal of input for societal impact[7]. As such it becomes a crucial question how to assess the impact of research for our modern life. To answer these questions, we may look for some hints from the latest trends of e-research and infrastructure initiatives.

In terms of user-driven business development, Europeana offers a good example as an e-infrastructure in the area of Digital Humanities and cultural heritage. In April 2016, it launched a beta version of Europeana Statistics Dashboard (Figure 1)[8]. It allows the users to examine the web traffic of Europeana from 2013 to the present. It is equipped with an easy-to-use interface to view the data from different angles. Although there are a few concerns about the "privacy" or "confidentiality" of web traffic information of the content providers of Europeana, it is most likely the transparency of the cultural heritage data aggregator overshadows the disadvantages. The author believes that it can trigger positive technical incentive to improve the metadata (and associated digital content) provided by thousands of content providers, as well as a political sentiment for the EU members to "compete" with each other. Similarly, the Archives Portal Europe, the aggregator of archives for Europeana also decided to share the statistics of their web portal[9]. Two big European initiatives seem to be moving forward to offer the transparency of their services and performances in order to meet the expectations of stakeholders for their credibility. Within the library sector, Farney and McHale (2013) pointed out that although very basic libraries are often asked to produce regular reports on web usage, including ones under the Association of Research Libraries in America. It should be equally highly rated that the Europeana regularly produces professional reports on the state of their operation, providing facts and figures of their finance and business in conjunction with Key Performance Indicators (KPIs)[10].

---

[7] For example, http://ec.europa.eu/research/participants/data/ref/h2020/grants_manual/pse/h2020-guide-pse_en.pdf (p13)
[8] http://statistics.europeana.eu/
[9] Personal communication with Wim van Dongen, the country manager coordinator of Archives Portal Europe Foundation, (2016-09-06).

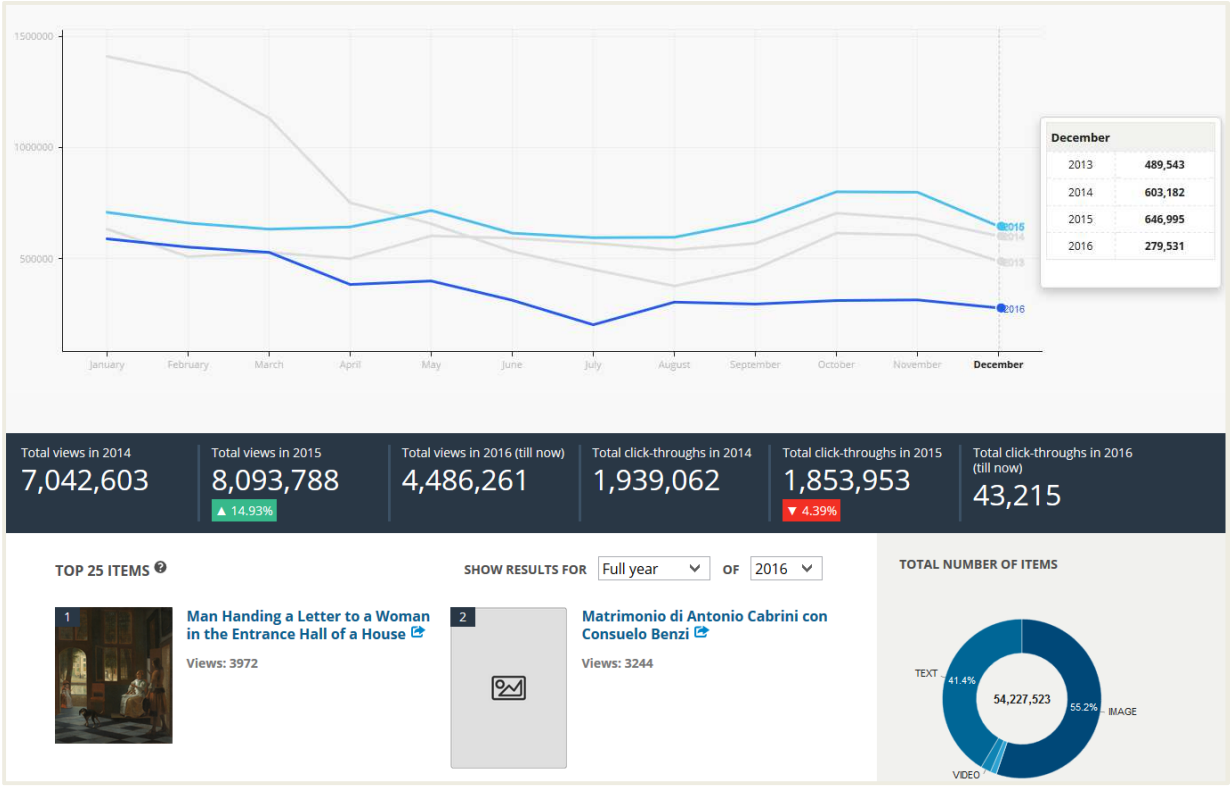

Figure 1. Europeana Statistics Dashboard[11]

In a broader context, a new form of research evaluation is considered to be urgently needed. This is especially discussed in Open Science initiatives. For example, the European Commission is keen to develop a methodology to measure research quality. Expert Group on Altmetrics clearly and rightly stated (Wilsdon 2016): "Wider use of quantitative indicators, and the emergence of Altmetrics, can be seen as part of the transition to a more accountable and transparent research system." In his thorough book, Stuart (2014) also discussed about different types of metrics and their history including bibliometrics, webometrics, and webmetrics for library and information professionals.

With those trends in mind, the analysis in this paper does not simply offer statistics and user evaluation, but also a hint of *Open Evaluation* of a research infrastructure in general. In other words, it is a challenge to answer such questions as how we can provide an open and fair development and services and how to measure their impact on the users and our society at large. Apparently the simple web statistics cannot be compared with such complex metrics, but it would try to serve as a small forward-thinking contribution to CLARIN as a research infrastructure, which would also be hoped to guide a way to Open Science and Innovation. Indeed, the most envisaged value of this paper is to raise awareness of the user-centric, transparent, and measurable approach for research and development within the CLARIN community.

## 3 Analysis between zero and one

CLARIN started to record the web traffic by Piwik[12] on June 23rd in 2014, although there are some variations of starting dates for different websites. The author decided to set a baseline analysis period from August 1st 2014 to July 31th 2016. That gives us reasonable information about the monthly and annual trend over 24 months, when recent critical development has been made.

As stated in the first section, CLARIN deploys a wide range of web applications. In this article, we will scrutinise VLO as one of the most prominent applications of all (Figure 2). There are good enough reasons for this choice: 1) A large amount of time has been spent on the development of VLO as well as the underlying concepts and services, most notably CMDI (Component MetaData Infrastructure)[13]. 2) It is a symbol of CLARIN's pan-European collaboration, created by people from many member states. 3) VLO contains a staggering number of records, currently over 900.000, about language resources and tools, which could have a big impact on our research-driven society. 4) Access numbers are one of the highest among CLARIN's applications. Consequently, the author would like to skip the analyses of other applications with regret, however leaving a possibility to work on them in the future. Although it is extremely valuable to analyse CALRIN's communication and outreach infrastructure including its main website[14] and social media such as Facebook[15] and Twitter[16], this paper concentrates on the technical user service, largely due to the limited access permission of the author.

[10] http://pro.europeana.eu/publications
[11] http://statistics.europeana.eu/europeana
[12] https://piwik.org/ In this paper, we used version 2.16.5.
[13] https://www.clarin.eu/content/component-metadata

### 3.1 Virtual Language Observatory overview

Figure 3 illustrates the two-year period of web traffic for VLO. At first glance, it is not what one might expect for an established research infrastructure. The total numbers of unique visits are 6612 in the Year 1 and 5260 in the Year 2. When translated into the average unique visitors, namely "visits per day", they are 18.2 and 14.4 respectively. The highest visits numbers are 45 and 39 respectively. Given the fact that CLARIN has more than nine years of history, there are 43 CLARIN centres in countries[17] (June 2017), about a hundred attendees participated in an annual conference (October 2016), and VLO contains more than 800,000 records (October 2016), it does seem that CLARIN can do much better. We must recognise the decreasing trend of 20.4 %, in spite of more promotion, visibility, and development in the Year 2. Although there is no particular benchmark at hand to compare, it is obvious that more effort can be made for marketing and dissemination, as well as the reconsideration of development priorities and strategies.

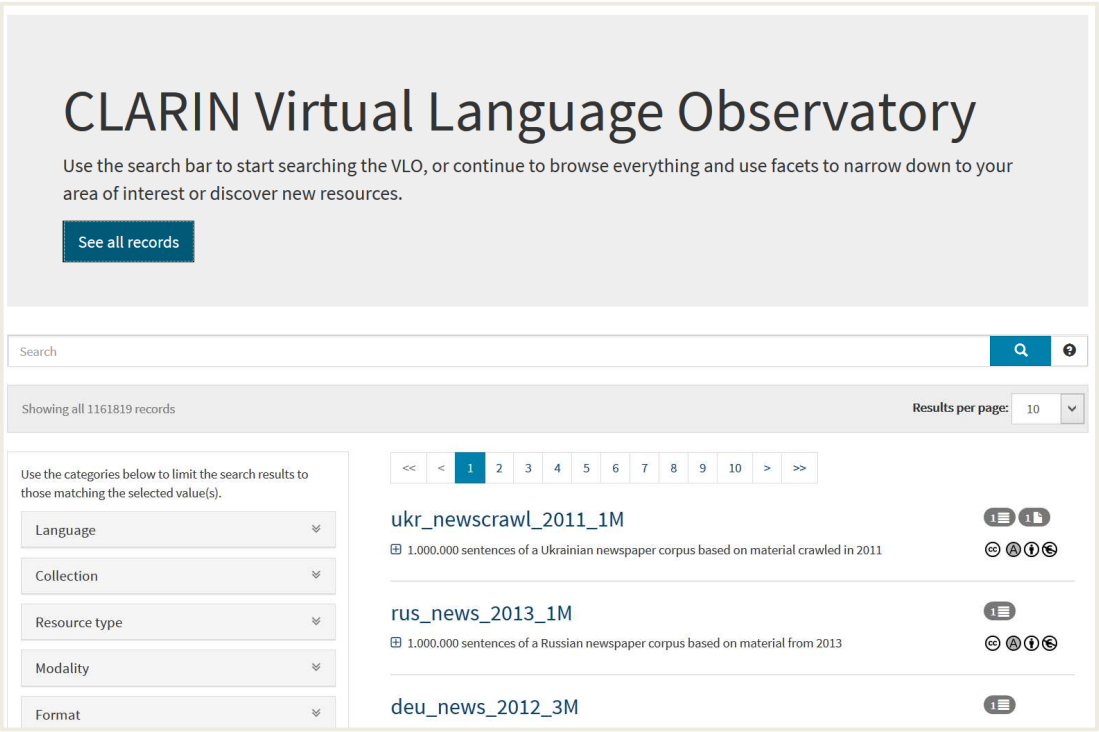

Figure 2. Current VLO home page

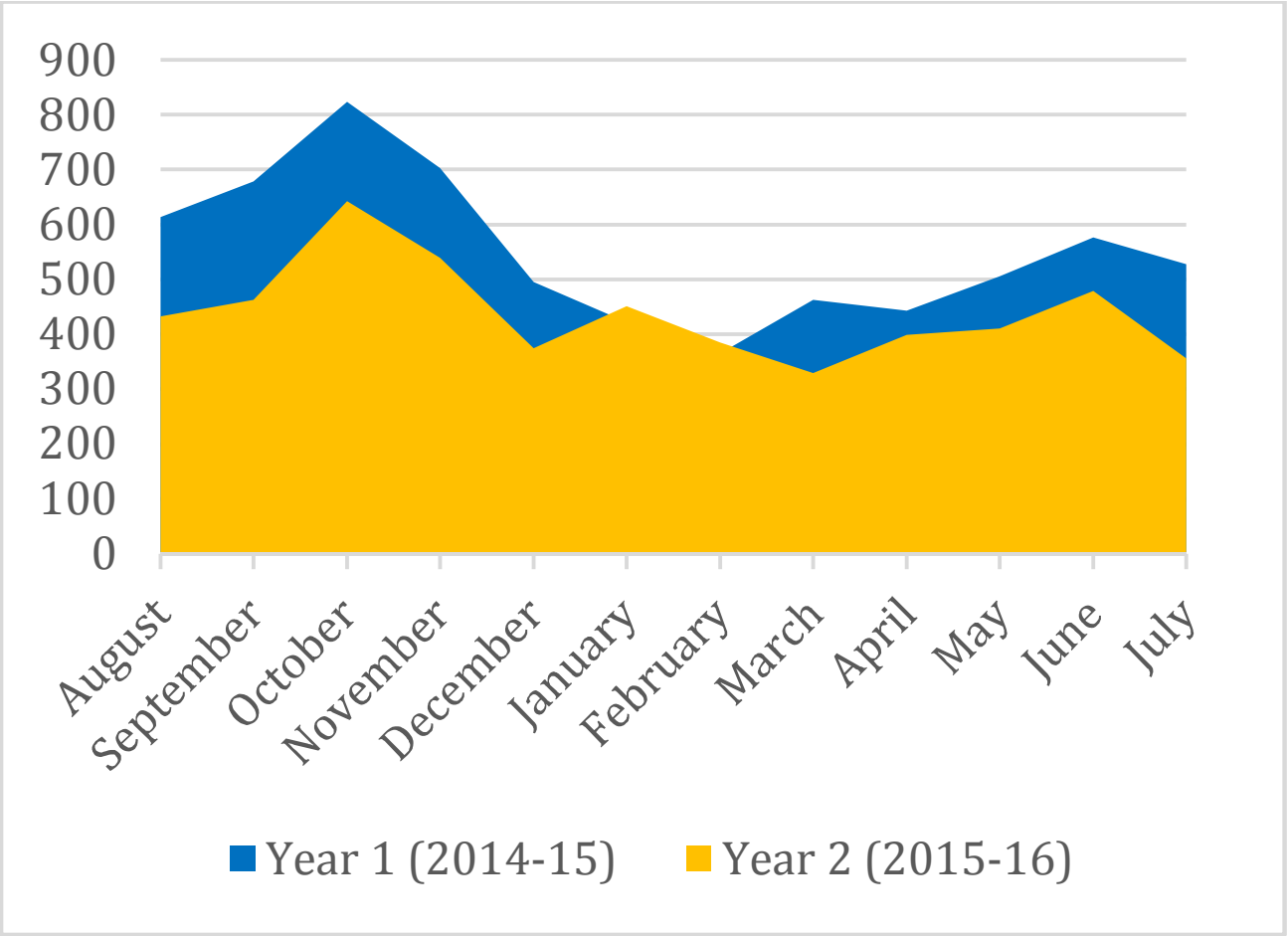

Figure 3. VLO unique visits for Year 1 (blue) and Year 2 (yellow)

There are hundreds of ways to play with the statistics, but we now look into the details of the overall picture. First of all, correlation with events is investigated. Figure 4 is a more zoomed-in version of Figure 3, showing the wave of access over time in the Year 1. It is common especially for small-scale non-profit workers-community websites that the access figures drop significantly during weekends. As the daily number of visitors is rather small, the lack of the main users (business-hour users) in weekends generates a very fluctuating graph which prevents data analysts from understanding overall trend. For this reason, the author pre-processed the original data that Piwik populates by excluding weekends, and created a new chart with better data-readability. Subsequently the new chart not only clearly reveals the slumps of Christmas and Easter holidays, but also a boost in the time of the CLARIN Annual Conference (CAC) 2014.

Year 2 is prepared in the same manner (Figure 5), but it focuses more on relevant events of CLARIN. When comparing with Figure 4, we notice that two figures provide a similar trend pattern, especially confirming the typical lack of access during European holidays. In addition, our common instinct within the CLARIN community is correct in that autumn brings more traffic when the biggest event of CLARIN normally takes place every year. In terms of other events, CLARIN was represented in a Digital Humanities conference in Kraków in July 2016, however, it did not really help the traffic increase for the VLO, which may raise an issue of the efficiency of promotion in the broad context of CLARIN's strategies for marketing and exploitation. More interestingly, the author pinpointed the timing of VLO releases. In Year 2, there were two minor releases and one version upgrade. Usually CLARIN announces a new release of VLO as a news blog entry, describing its technical improvements. In contrast to the great effort of the developers and others, those releases and their annoucements have very little impact on the web access. Version 3.4 and 4.0 were below or around the average, while version 3.3 can be interpretted as a part of the "domino effect" of the autumn boom. The detail examination of access over time is supposed to be very useful to check what happened before and after known events such as a new release of VLO, an addition of a new CLARIN member state, and a promotion of CLARIN at a conference. We can, therefore, plan the marketing and evaluate the impact more objectively.

---

[14] https://www.clarin.eu/
[15] https://www.facebook.com/ClarinEric/
[16] https://twitter.com/clarineric
[17] https://www.clarin.eu/content/overview-clarin-centres

### 3.2 Google Indexing

One last, but not least, pointer of Figure 5 is Google indexing. In 2015, the author recommended the VLO technical team to index VLO for Google in order to dramatically increase the web traffic of VLO. This proposal was accepted, followed by the generation of site maps by the Austrian Center for Digital Humanities (ACDH) in January 2016. Consequently we submitted them to Google in February. The sitemaps allow the Google indexer to crawl the content of VLO and the VLO records will be searchable by Google. However, the consequence was disappointing. Only 10,098 records/pages of VLO were indexed out of 881,338 described in the sitemap. In total, 37,172 were indexed out of 881,334 records by October 10th 2016[18]. It is also apparent that there is almost no impact on the web traffic, therefore, invisible in Figure 5. The author witnessed four or five times rise of the traffic per year in case of the Archives Portal Europe, when Google indexing is involved. The case is still under investigation by the CLARIN's technical team, however, there are two major obstacles. Firstly, Google indexer is a black box, and it is extremely difficult to know what it favours. The best effort was made by our team to follow the instructions and guidelines of Google to generate the sitemap. Secondly, VLO is dynamically generated and due to its infrastructural set-up, the records are harvested and updated regularly without implementing persistent identifiers. Such volatility may make Google find us unattractive. In the meanwhile, there are also some spikes in the figure, but the reasons of those are still an open question.

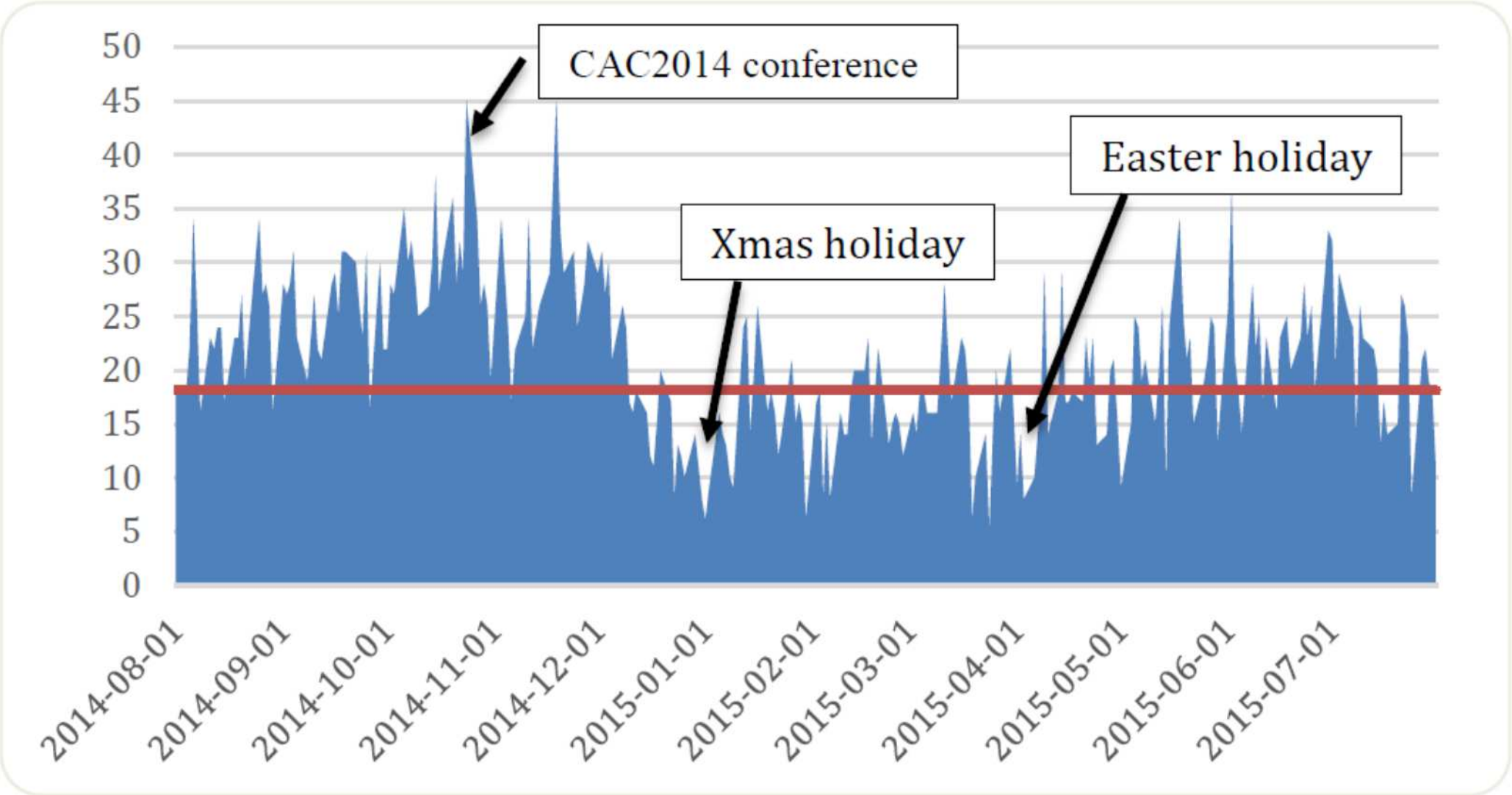

Figure 4. VLO unique visits for Year 1 except weekends (the red line is the average 18.2)

[18] The slight gap between the number of the sitemaps and records also imply the volatility of VLO and the difficulties in accurately counting records which regularly change over time.

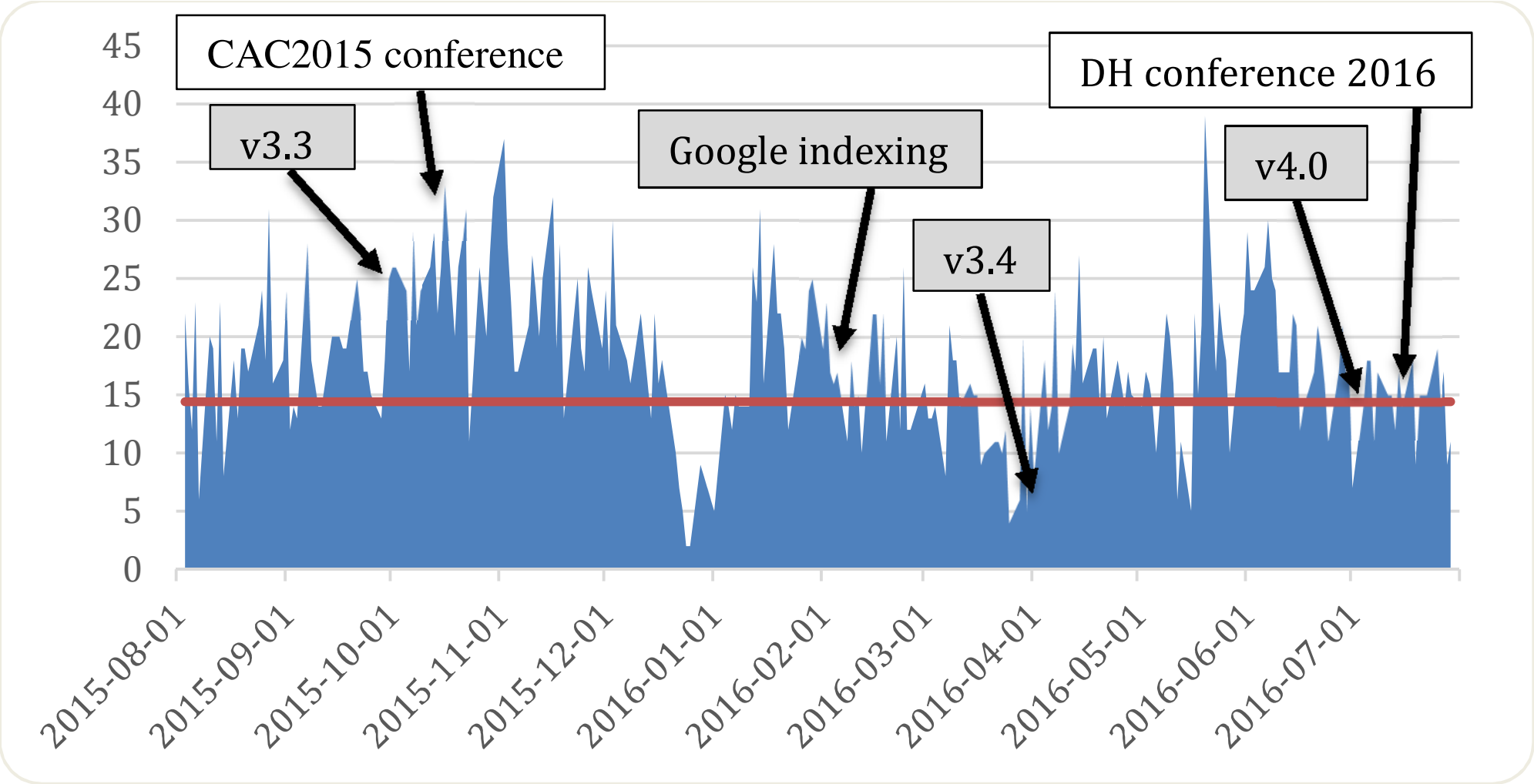

Figure 5. VLO unique visits for Year 2 except weekends (the red line is the average 14.4)

### 3.3 Profile of the users –geographical distribution

The next stage of our investigation is to try to figure out who the users are. One of the obvious questions is where they are from. Place is one of the 4Ps in marketing mix[19]. It is an important factor of marketing strategies and an analysis of the geographical attributes of users is a must. In VLO, the top countries of access are Germany (2830 visits, 31.6%), the Netherlands (1331, 14.9%), Austria (772, 8.1%), and the United Kingdom (628, 7.0%). Figure 6 translates them into a European map. The domination of German access is clear and it is very tempting to stop here.

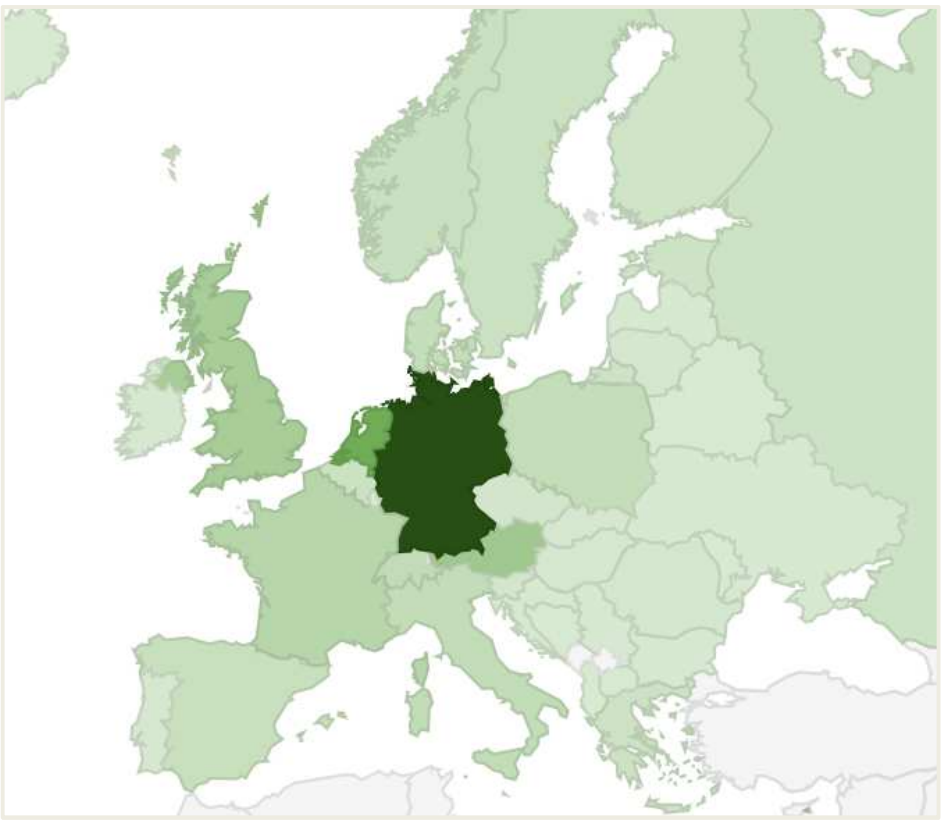

Figure 6. Geographical distribution of the users in Europe[20]

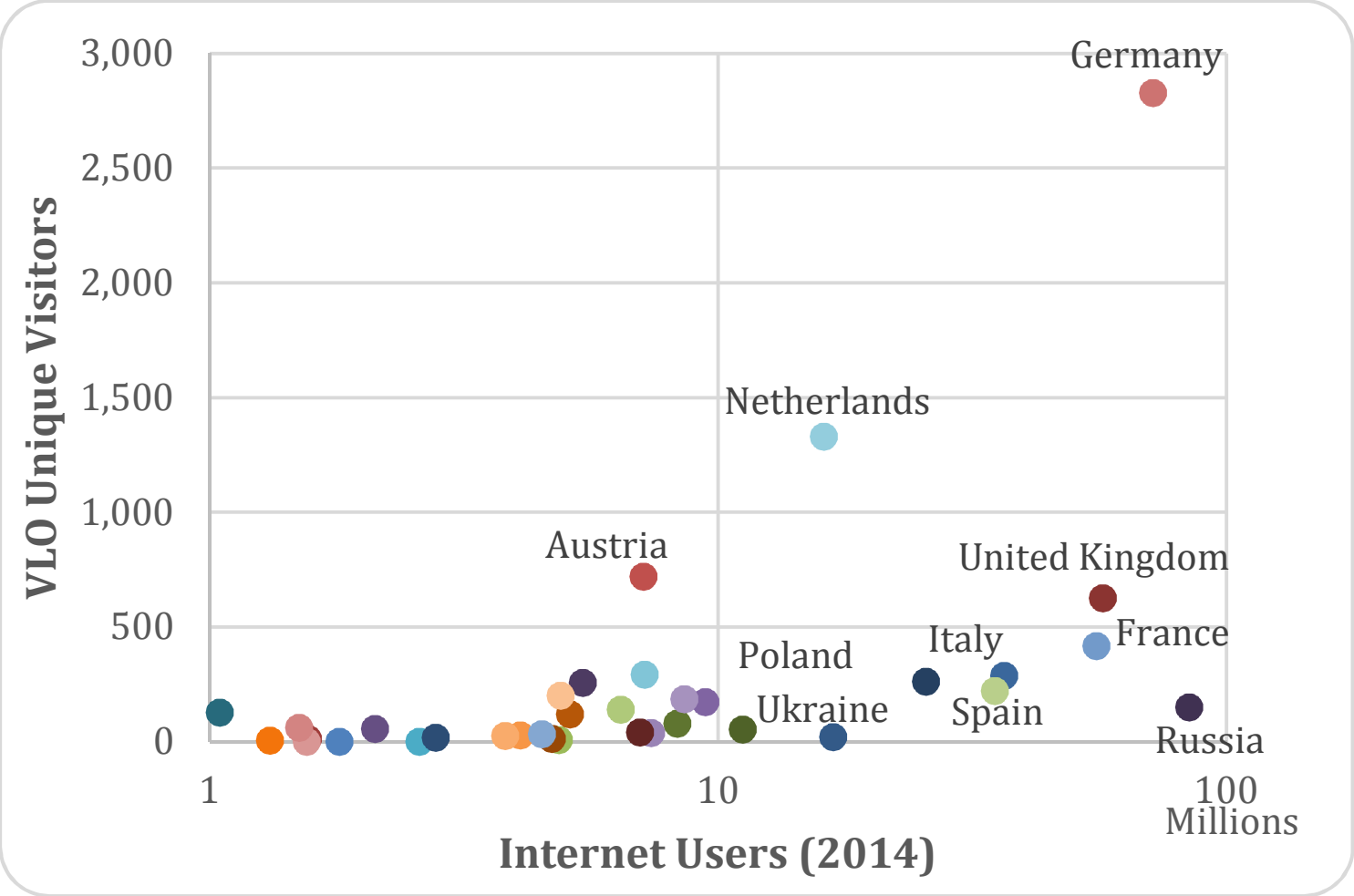

Figure 7. VLO unique visitors and internet users in Europe

However, it does not take a minute to realise that Germany has more population than a small country like Malta, therefore, it is often the case that Germany wins in Europe. To rectify the result, the numbers are contrasted with the statistics of internet users per country provided by Wikipedia (Figure 7). In this chart, the horizontal axis takes internet users and the vertical axis for VLO unique visitors, illustrating the relations between the internet users and the VLO users. Germany is still exceptional, while Russia underperforms. The UK and France do not demonstrate their strengths, but the Netherlands and Austria remain in relatively high positions. However, visualisation is sometimes misleading for the right interpretation. Figure 7 only focuses on the relativity of countries, therefore, it is still not totally convincing for our purpose of analysis. Now, if we simply present the ratio of VLO users per internet population, we can find the countries in which VLO are more popular than others. Figure 8 confirms that 0.012 percent of internet users in Estonia accessed VLO, followed by Austria (0.010%) and the Netherlands (0.008%). There are new emerging countries previously unknown such as Denmark Norway, Slovenia, and Switzerland which are above 0.004% threshold. This example signifies the importance of adequate data normalisation and visualisation.

---

[19] https://en.wikipedia.org/wiki/Marketing_mix

[20] The default visitor map in Piwik counts visits (not visitors) and this map is created by Google Sheets

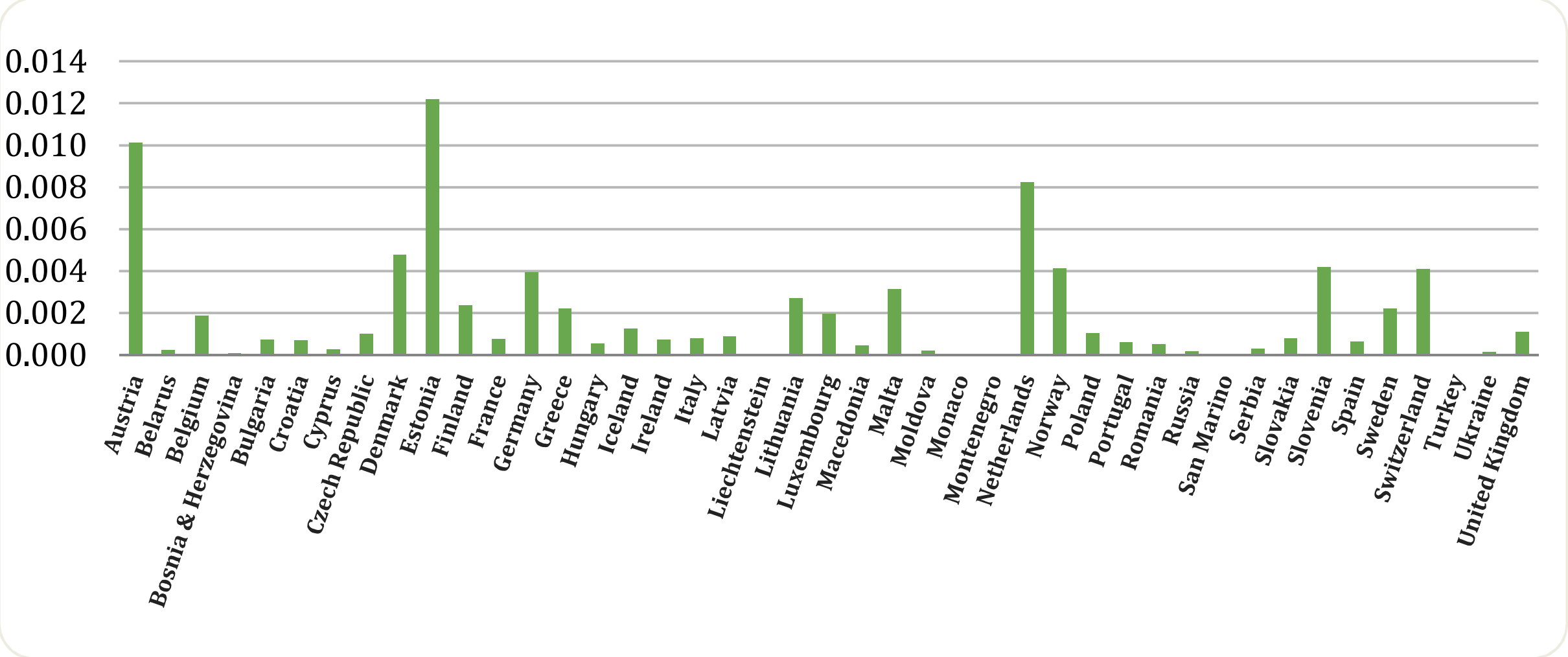

Figure 8. Percentage of the VLO users in the internet population in Europe

Similarly, Piwik does not have a facility to visualise the data for all basic needs of the users. For example, plotting most accessed cities on a map cannot be done automatically. The list of European cities with most access was needed to be converted into the list with coordinates. With its easy-to-use Datasheet Editor[21] developed by DARIAH-DE, cities names are automatically detected and geo-coordinates were assigned with a slight possibility of errors misidentifying a few of them as US cities. Then, a coupled application, Geo Browser, seamlessly creates a distribution map (Figure 9). This kind of map is very helpful to build strategies for marketing. For instance, by comparing it with a map of CLARIN centres (Figure 10), one can observe similarities and differences. This could lead to a question whether CLARIN should intensify its activities in the place where VLO is popular, and/or explore a new area to find new users. That would be also a useful addition to such a report as the analysis of the national roadmap for research infrastructures for non-CLARIN countries (Maegaard and Olsen 2016).

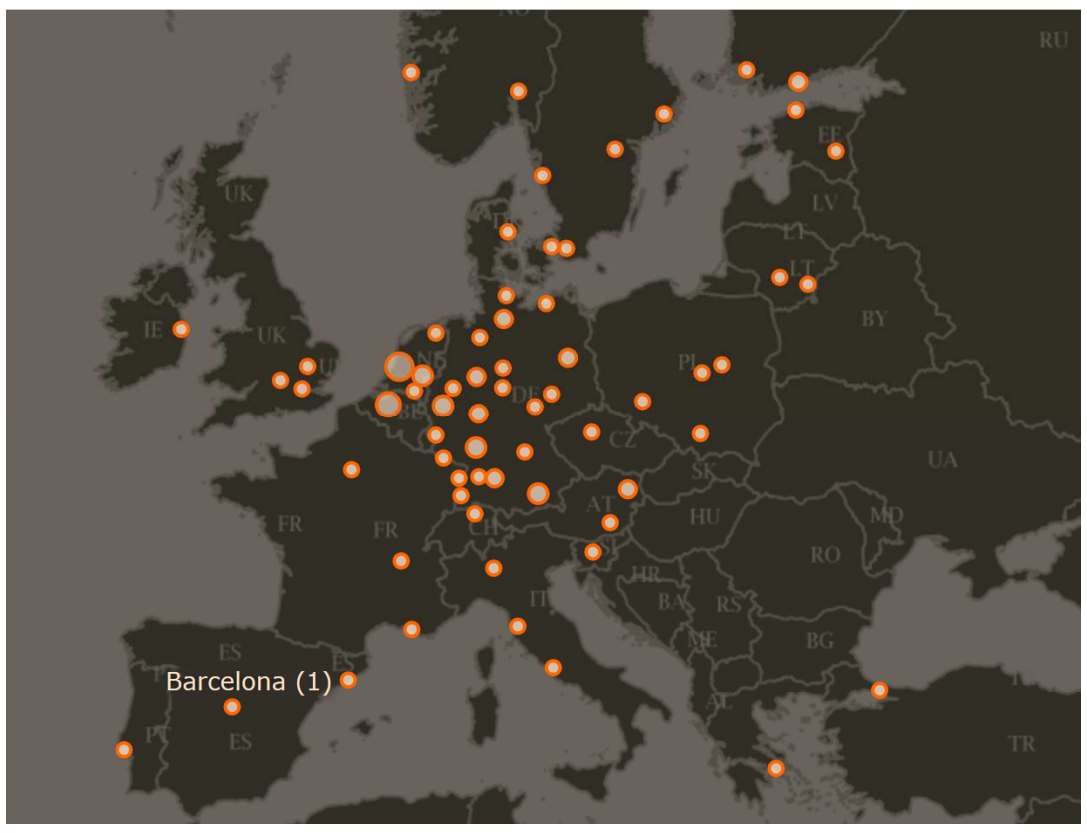

Figure 9. Top 100 access from European cities (by Geobrowser)

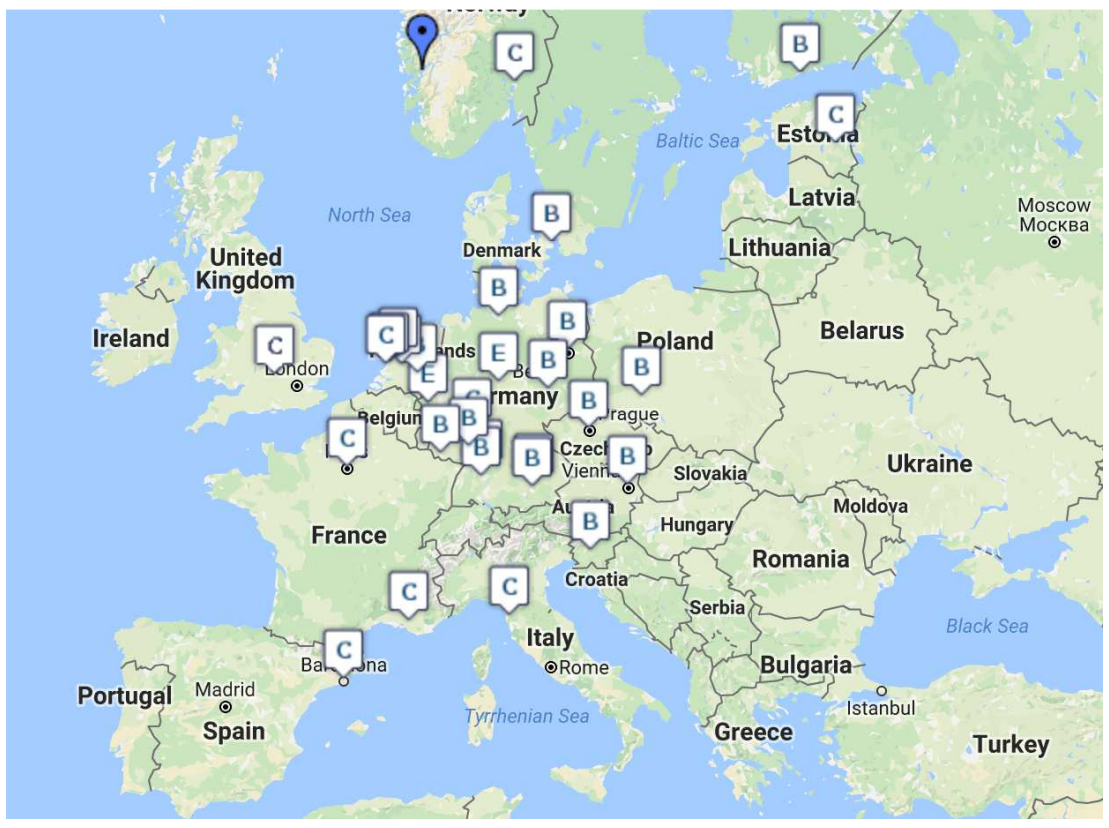

Figure 10. Map of CLARIN centres[22]

[21] https://de.dariah.eu/geobrowser
[22] https://www.clarin.eu/content/overview-clarin-centres

Europe is our primary concern as most of the CLARIN activities are held within. However, globalisation is something not to be ignored. 22 % of the VLO users originate from outside Europe (Figure 11), which implies that there are a number of international users who do not yet belong to the CLARIN consortium. While our usual suspect of European cities such as Vienna (1st in the city ranking, 953 visits), Hamburg and Nijmegen (both 2nd, 404) and Tübingen (3rd, 292) catch our eyes, there are a number of key US cities in the score sheet, including Ashburn, Virginia and New York (both 22nd, 62) and San Matteo, California (32nd, 42). Together with Lima (37th, 31) and Mexico Distrito Federal (43th, 28), they sneak in the top 50 cities among all other European place names.

There are many other data views to scrutinise the data from a location perspective. For instance, Figure 12 is a bar chart of visits from Austria (1179) and the United States (1252) which is put in the time context. It is clear that the access from the United States declines as time goes by, in sync with the general trend we observed in Figure 3, whereas Austria goes contrary, becoming a highly active country around mid of 2015. It is probably safe to say that the establishment of the Austrian Center for Digital Humanities (ACDH) is the primary reason for this dramatic shift. It started to employ people in spring and summer 2015 and some of them have heavily worked for CLARIN since then. As a result, the numbers are sky rocketed. As the United States is not the official consortium member of CLARIN, this result manifests the impact of CLARIN's internal use. The marketing implication is as follows. The expansion of the consortium is a great strategy to promote CLARIN and convert the external community into the internal one to gain more access to VLO, but it is more likely that the inflation will stop at some point when most of European countries joined in. From that point on, if CLARIN is eager to stretch their use-base, it will need more inbound access from outside the consortium, which may be much harder to achieve and sustain.

Although we can share a preliminary analysis and indicate the potential use of Piwik data like this, this paper cannot offer all the cross-matrix of city names and visit duration and bounce rate etc (also in relation to time scale), it is certainly interesting to have a close look of specific geographical access habits.

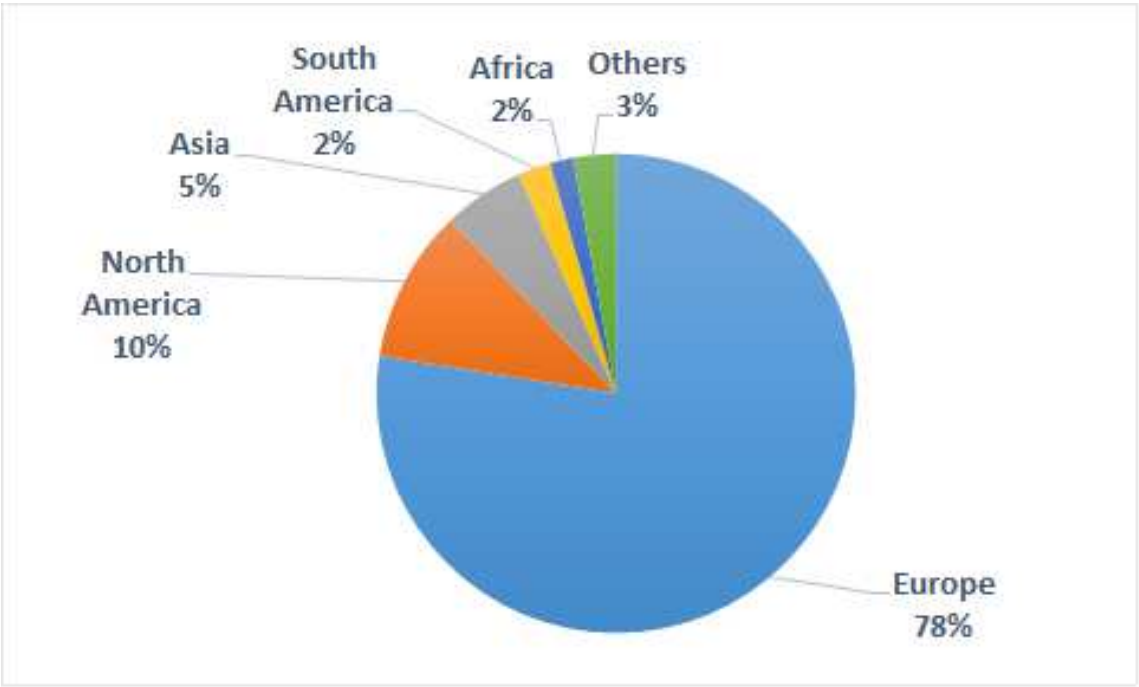

Figure 11. World access demography

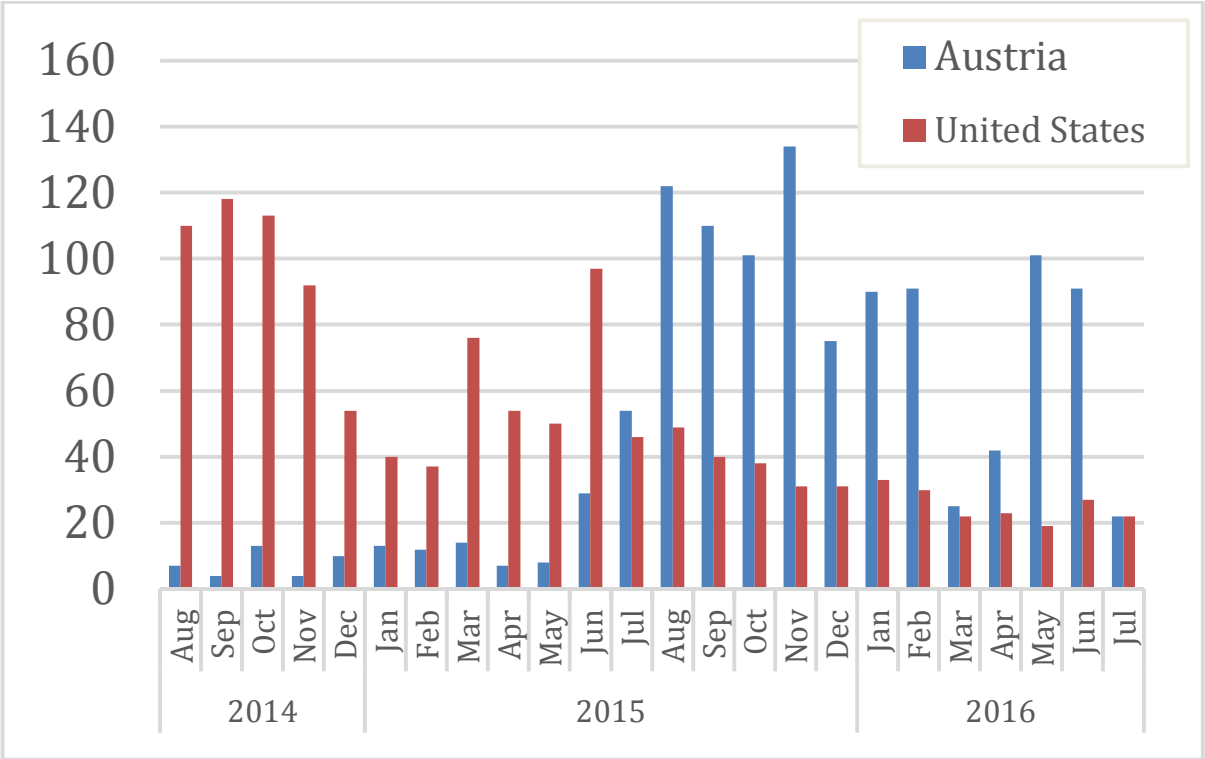

Figure 12. Visits from Austria and the United States

### 3.4 Visit duration and visit frequency

Other basic statistics are the visit duration and frequency. They would indicate the level of user engagement. Most of the VLO users spent less than 10 seconds (Figure 13), which may imply that they often exit a web page without examining the content as soon as they find it useless, accomplish a technical testing, or for other similar reasons. However, the users spent the average of 4 minutes 18 seconds, suggesting that there are minority users who stay in a website for much longer time. In fact, 3% of the users visited the website more than 30 minutes. One of the obvious reasons for this user divide is the internal users of the CLARIN community. In order to identify the bias of the internal users, the data was filtered by a CLARIN partner institution –ÖAW (the Austrian Academy of Sciences) under which the ACDH is founded (Figure 14). This is easily executed by specifying visitors' IP address in one of the filtering options of Piwik. The result convincingly shows the heavy usage of VLO by the ÖAW. More than half of the ÖAW users stay in VLO more than one minutes, while the average dura-tion is 8 minutes 42 minutes. The comparison between the two user groups suggests the need of close and careful examination of the basic data automatically generated and presented as a quick-result in Piwik.

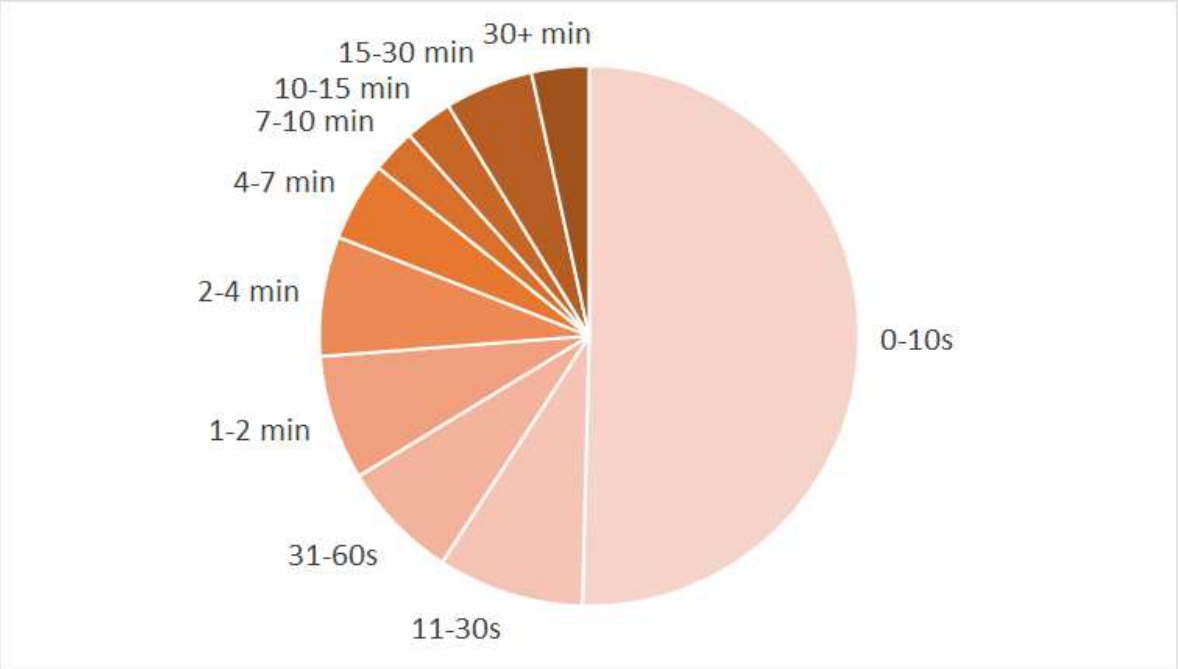

Figure 13. Visit duration for all users

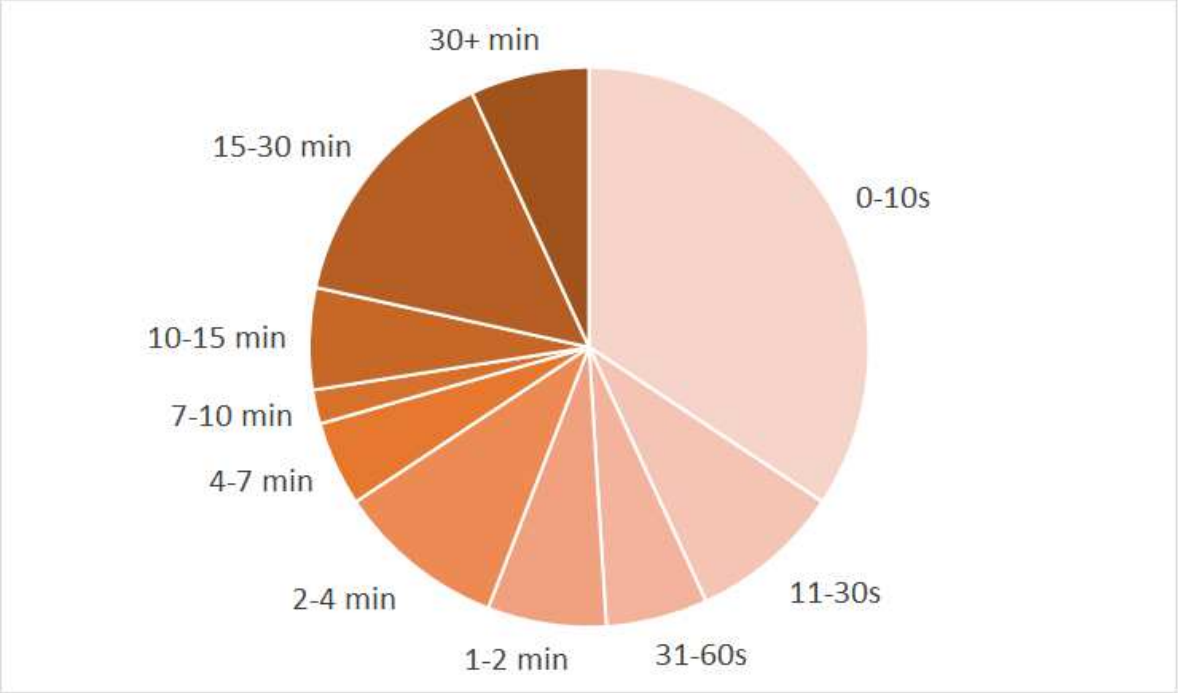

Figure 14. Visit duration for the users from ÖAW

Visit frequency charts require the same attention. 59% of all the users are the first timers, whereas there are visitors who visit the website more than regularly (Figure 15). The ÖAW again is in a completely different scene. About a half of the visitors accessed the website more than eight times, and the repeater group of 51-100 times is the most prominent of the pie chart occupying 38% (Figure 16). It is apparent that the ÖAW's user behaviour is totally different from that of the overall users. This outcome also signals a need for caution.

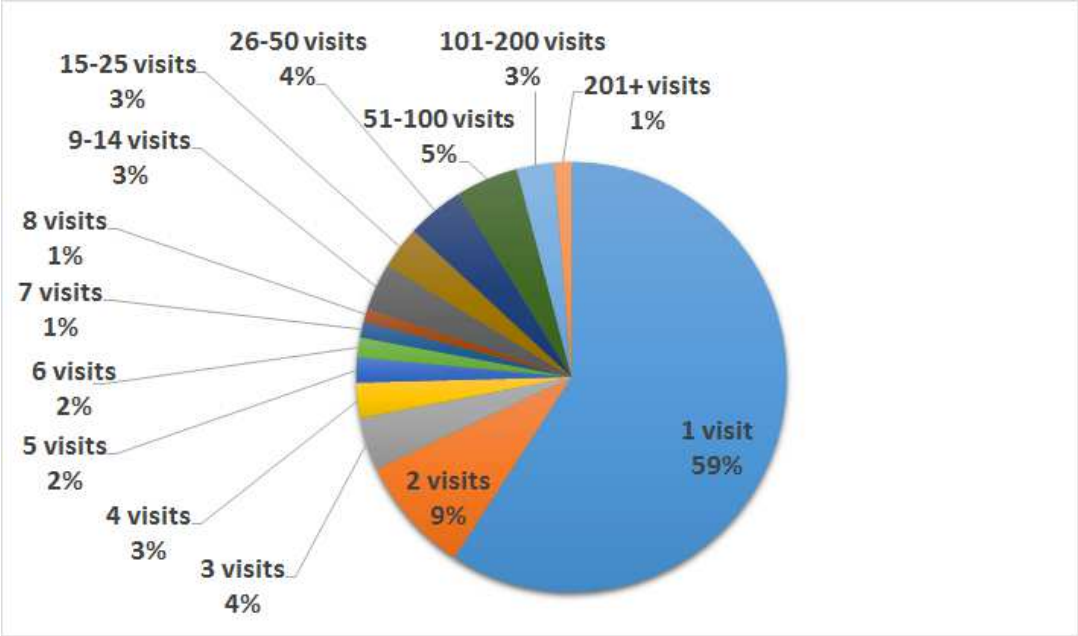

Figure 15. Visit frequency for all users

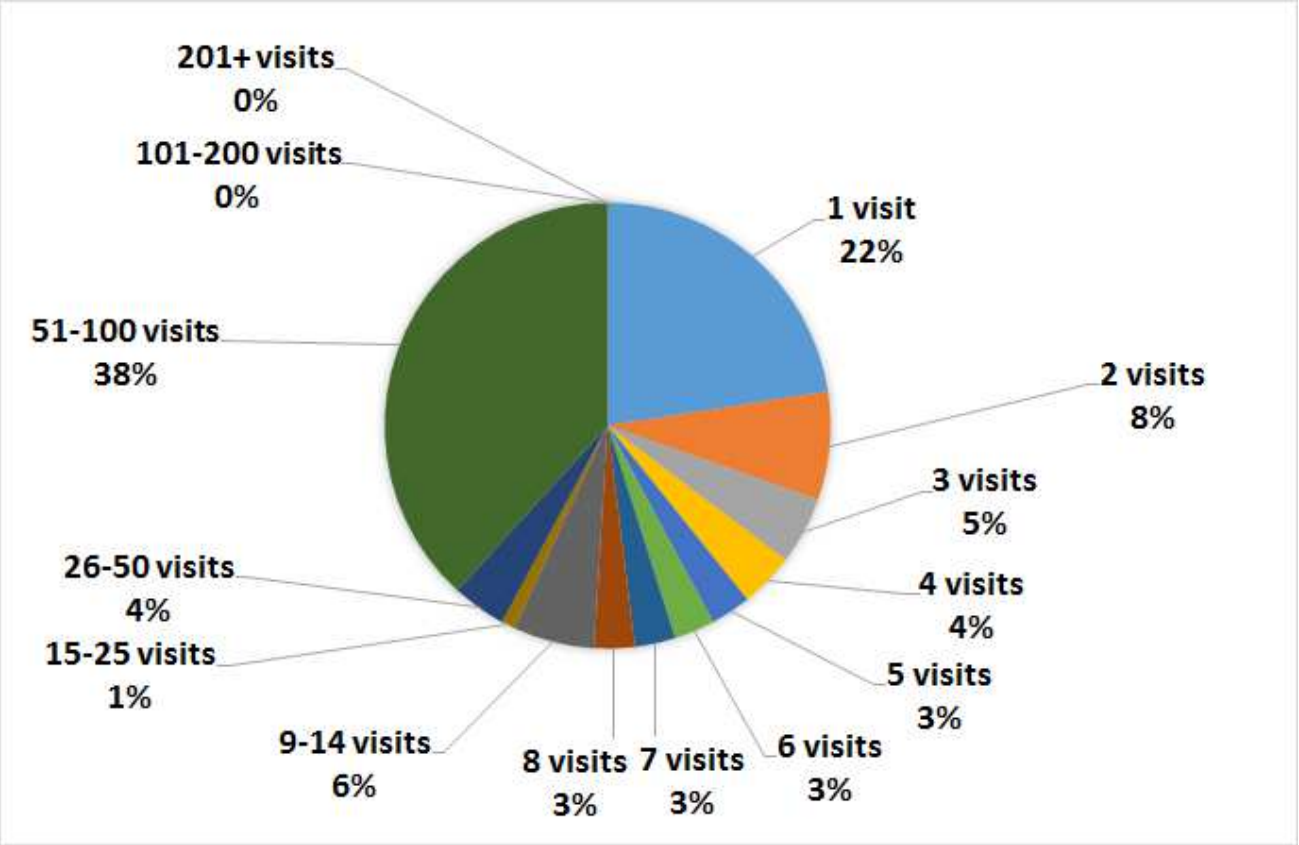

Figure 16. Visit frequency for the users from ÖAW

## 3.5 Search keyword frequency, language, and social media referrers

In order to understand what the users do and why they visit the website, other metrics are required. The keyword frequency list provides us a useful insight into what the users look for. In total, 3506 keywords are recoded, in which potential duplicates may exist because of normalisation problems. Figure 17 is the tag cloud of the search keywords used in the VLO (section 4 will briefly describe how to generate tag clouds from Piwik).

The first discovery is that English is predominantly used for the search keywords. This may not be a surprise, but reasons should be sought. It may be simply because the VLO interface and content target international users (successfully, or perhaps too much). It may a result of the growing number of internationally-minded researchers, or the reconfirmation of globalisation and diversity crisis by means of English as lingua franca. Further, it may be the consequence of a high volume of access by the internationally-trained CLARIN researchers, without managing to reach broader audiences. Whether those reasons are interpreted positively or negatively, the situation should be kept monitored.

The most common words are “geco” (102), “hzsk” (89), “german” (50), “corpus” (45) and “dutch” (45). There is a tendency toward language names (especially European ones) and this result is interesting, because VLO offers language code facet to filter the search results. More interestingly, King et al. (2016) suggested that its facet coverage is between 30 and 40%, meaning about two thirds of the records are not visible when filtering by a facet. This might explain why the users may prefer to use language names as search keywords, but it is too early to draw a conclusion, as the problem of facet coverage was not informed thus unnoticeable in the VLO interface at the time of publication[23].

In contrast, as far as the users from Vienna (where the ÖAW is located) are concerned, language names are not the primary languages of searching (Figure 18). Keywords are mixed and specific, and perhaps more oriented towards technical terms such as the names of services and researchers' areas of interest. It is also noted that language takes an influential role for VLO search experience, especially when the metadata is described in the language(s) of resource providers.

Aligning with the section 3.4, this is another evidence of the need of understanding different types of users and creating marketing and development strategies for user segmentation. There is no doubt that the analysis of search keywords could contribute to (re)write user requirements which are not yet known and/or not found from other methods of user evaluation.

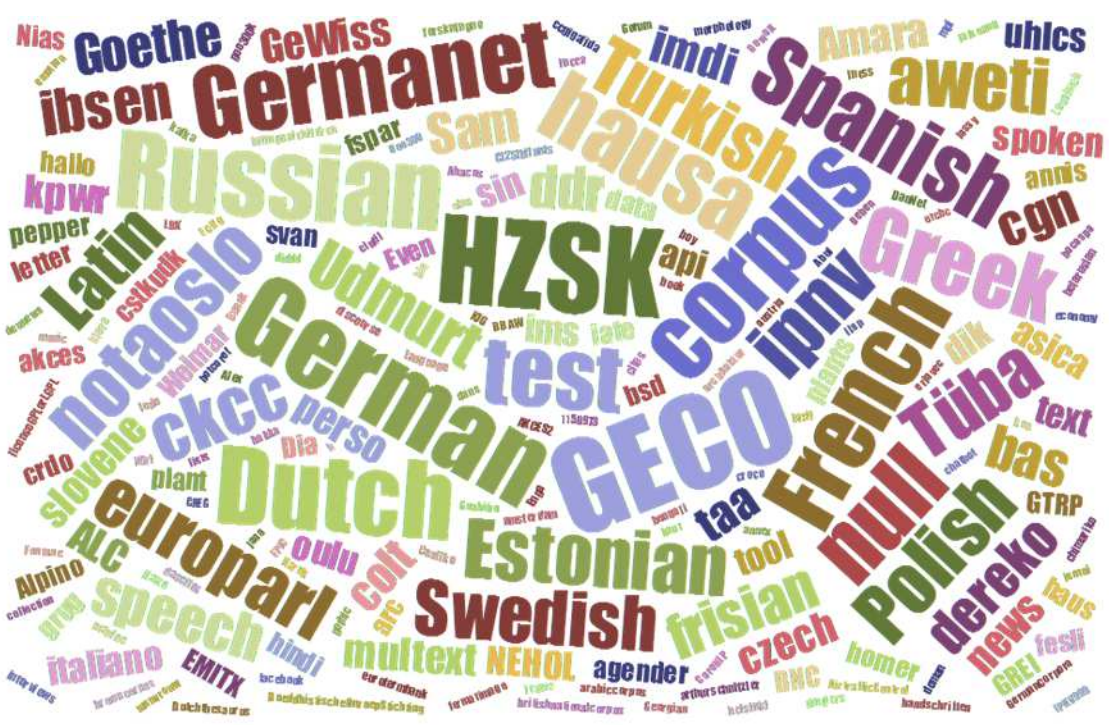

Figure 17. Keyword frequency for all users (tag cloud)

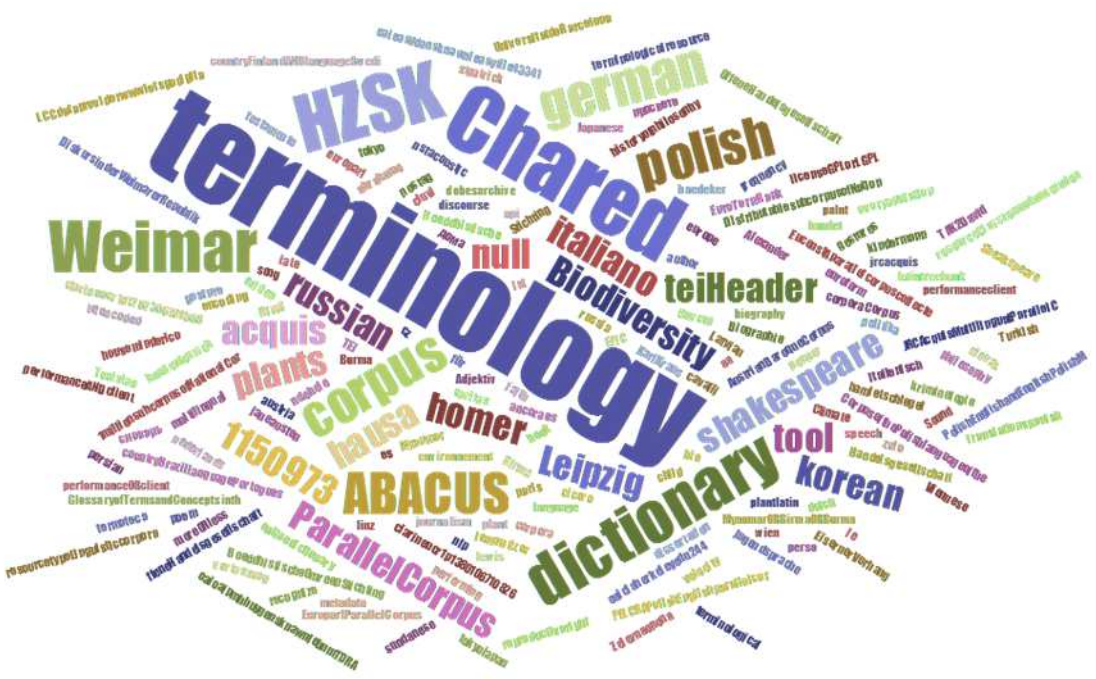

Figure 18. Keyword frequency for the users from Vienna (tag cloud)

Relevant to this language discussion, Piwik collects some more language information. Figure 19 represents the browser languages used by the users. Half of the users use a variant of English, while German is the only real contender (22%) far ahead of other European languages. Except German, it turns out that the VLO users are as international as we saw in the tag cloud. To some extent this is also confirmed by the fact that a surprisingly high percentage of ÖAW users sets US English for their browser language (Figure 20). It seems that they are extremely international and do not use (Austrian) German as their choice of language. Latvian (2%) is totally unexpected, casting a little doubt about the accuracy of Piwik recording, although it is not possible to deny that Latvian users exist in the ÖAW. Provided that the ACDH would be the most obvious user group of the ÖAW, implication is that we as CLARIN researchers are too international to use their mother tongue. If a wild guess is permitted, this argument could be extended to the point where Figure 15 and 16 can be interpreted more differently than we initially thought. The 38 % of the users in Figure 16 may largely overlap the CLARIN's highly international researchers, whereas German, as well as small language groups (Dutch, French, Spanish, and Italians etc) may represent the majority who accesses much less frequently. They are more likely to be non-CLARIN users and/or novice users (eg non-technical, or non-international, or non-linguistic humanities researchers), in other words, the users that CLARIN may want to explore intensively to expand the use of CLARIN infrastructure. It is one of the benefits of Piwik that it can record the chronological statistics of particular segments, be it a segment of a language or search keyword, so that we can follow the user segments over time to potentially narrow down the specific user groups. In any case, more scrutiny on the data would bring us more evidence.

[23] The VLO environment is changing constantly in the course of product evolution.

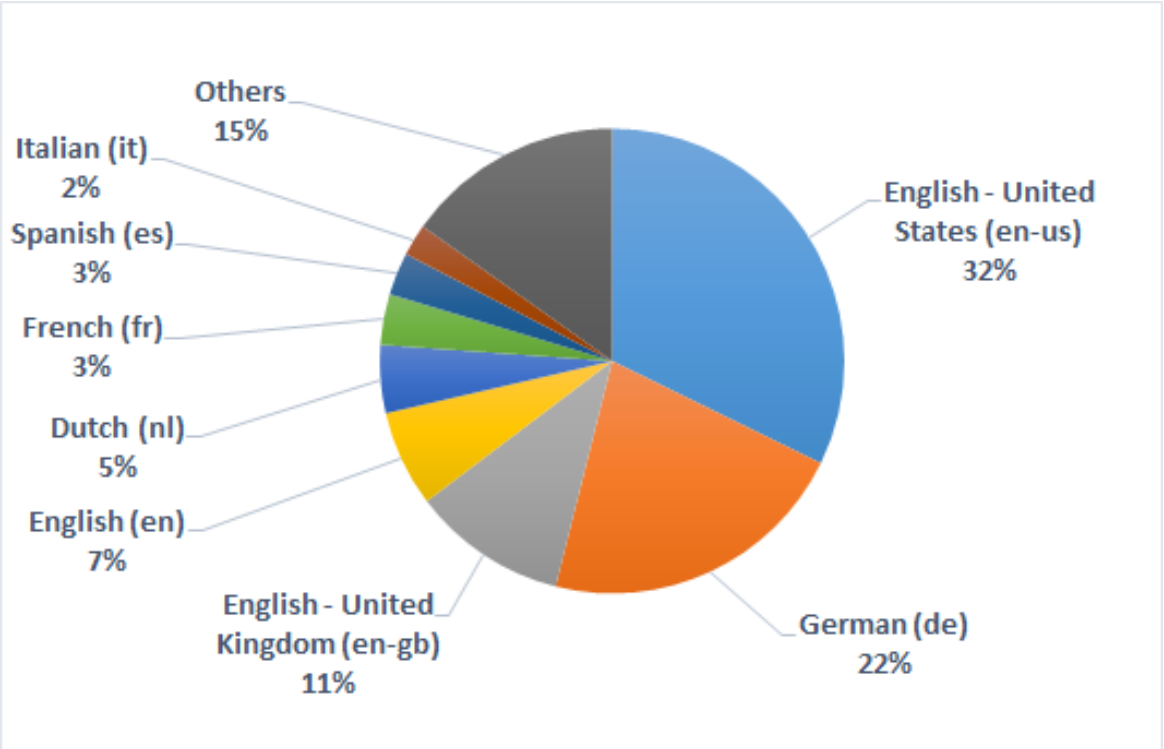

Figure 19. Browser language for all users

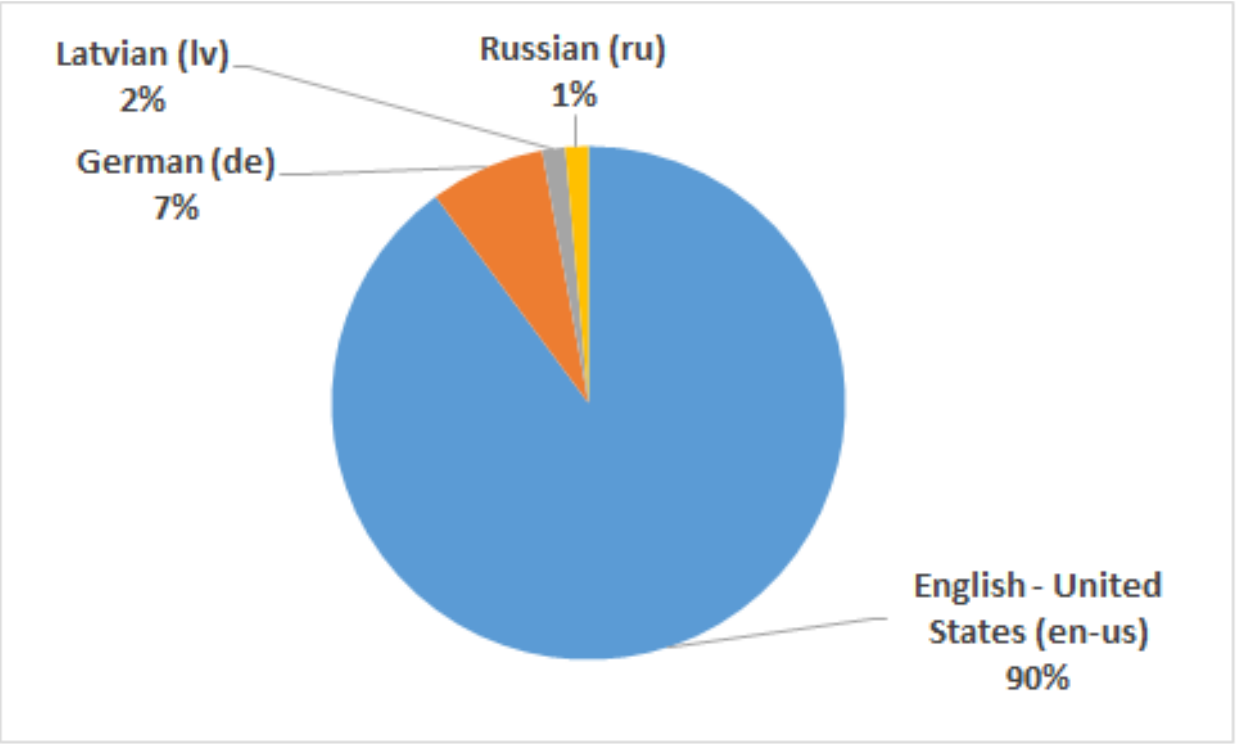

Figure 20. Browser language for the users from ÖAW

Another fascinating observation can be made when comparing the search keywords of the VLO with the keywords of web search engines to be used to find VLO and/or its content within. However, it turns out less useful, because 88.7% of the data is classified as "keyword not defined" (Figure 21). In terms of the science of web analytics (section 4.3), this is a good example of being best not to compare due to the proportion of available data. Similarly, compared to the website referrers, social media referrers are too few to interpret (Figure 22). If those situations would change in future, the comparison of the statistics may produce interesting results.

**Keywords**

| KEYWORD | VISITS |
|---|---|
| Keyword not defined | 3,424 |
| clarin vlo | 7 |
| https://catalog.clarin.eu/ | 4 |
| paradisec | 4 |
| rongmei bible | 4 |
| vlohausa | 4 |
| hzsk | 3 |

Figure 21. Most visits are "(search engine) keywords not defined".

**Websites**

| WEBSITE | VISITS |
|---|---|
| www.clarin.eu | 2,047 |
| clarin.eu | 762 |
| clarin07.ims.uni-stuttgart.de | 104 |
| www.ims.uni-stuttgart.de | 92 |

**Social Networks**

| SOCIAL NETWORK | VISITS |
|---|---|
| Twitter | 20 |
| Facebook | 6 |
| GitHub | 2 |
| Vkontakte | 1 |

Figure 22. Very low visits for websites referrers and social media

### 3.6 Change of user behaviour

The Pivot Table of EXCEL is a powerful tool to examine data from various perspectives. The author experimented it in order to find any interesting outcomes. In the course of two years, somehow searching gains momentum (line chart), while download loses it (bar chart)(Figure 23). Taking the section 4.3 into account, here we introduce a hypothesis building process. The motives of the change of the user behaviours are not yet known, but it seems that the summer in 2015 is a key moment of this clear cut. It is around the time of the release of the version 3.2 (early July), or less likely 3.3 (end of September). It has to be also remembered that there is an overall dropping trend of unique visitors. Although not as drastic, it corresponds to the decrease of downloads. With our limited information, it is not easy to identify the exact cause of this issue, but the release reports of VLO developers may have some hints though two releases are rather minor updates. On the download side, the version 3.2 introduced a pagination for a list of resources to download[24]. In addition, availability facet was implement-ed to display license information including "free" and "free for academic use". On the search side, results ranking got defined by relevance. In the version 3.3, advanced queries were deployed so that the users can use regular expressions such as AND, OR, and NOT to combine search keywords[25]. The ranking algorithm was further fine-tuned to prioritise primary metadata fields in order to meet the expectations of the users. Whatever the possible reasons of the change of the user behaviours, more analysis of this phenomenon and continuous monitoring are needed to draw a better conclusion.

[24] https://www.clarin.eu/blog/updated-vlo-brings-better-ranking

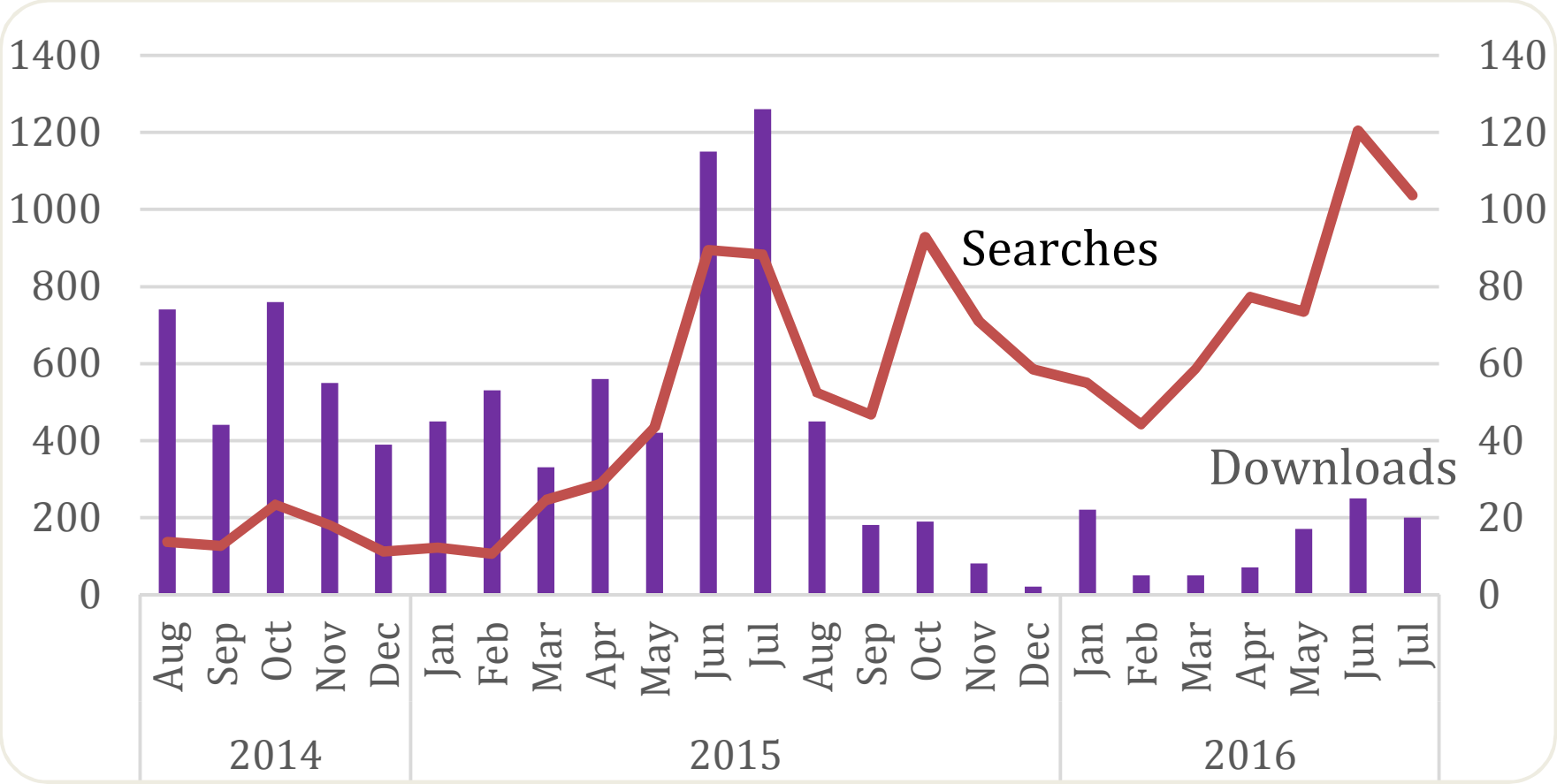

Figure 23. Number of searches (line) and unique downloads (bar)

## 4 Playing a fair game? –methodological issues of web analytics

This section aims to discuss the procedural and methodological aspects of web analytics in general. It is intended to focus on "the science of our experience" on analyses, hypotheses and interpretations of web traffic, which is relatively untouched in the studies of web analysis. The section is divided into two parts. The first part will try to open and analyse the black box of web analytics. The second half will discuss the techniques and tips used for the analysis in order to document the research experience and share it with other analysts of the future.

### 4.1 Opening the black box

One of the main questions of web analytics tools is how trustful they are and how to adequately analyse and interpret the statistics they produce. It is also important to know if the statistics of different websites are comparable. Although there are official documentations of web analytics about the counting methods of user access[26], web analytics applications are often a black box and many marketers blindly use them without knowing exactly what is recorded within. To this end, one experiment is made to understand how Piwik keeps tracks of the web users. Its visitor log and visitor profile function enable the author to monitor every move of a particular user (Figure 24). This time a colleague of the author was chosen to conduct a simple experiment. An arbitrary protocol was set to execute a series of tasks and the visitor log is analysed to assess what is recorded:

1. Go to www.google.at with Firefox
2. Search exact phase "VLO CLARIN Bulgarian"
3. Click the second result of VLO (Bulgarian Alphabet, Phonemes, and Spelling)
4. Click "VLO" link at the top to go to the VLO Home Page
5. Search exact phase "hungarian"
6. After the result page is shown, kill the browser tab

This trial revealed that the basic information such as the timestamp, country (plus the region and city), operating system, and browser is correct, while the actions taken are perhaps not the same as expected (Figure 25). Although Google was detected as the entry point, the search term was not identified and classified as “keyword not defined”. Indeed, this is a reconfirmation of Figure 21. The click of a specific record was correct, and the following path to the home page was accurately retrieved, so as the last trace of his activity: search keyword “hungarian”. The second experiment was carried out in the author’s environment (Internet Explorer) with a longer protocol:

1. Go to https://www.clarin.eu/content/virtual-language-observatory-vlo
2. Click the link to VLO
3. Search exact phase “Latin”
4. Click the last result of the first result sets (“LangBank Latin Corpus”)
5. Click Resources tab
6. Click “caeser1.cha” file to download
7. Click back button and click Availability tab
8. Click “Back to the result” link at the top
9. Click page 2 of the result sets
10. Click the third result (“Latin Lemmatizer On Line”)
11. Click Resources tab
12. Click “latmorphwebapp”
13. Kill the browser tab
14. Repeat the task from 1 to 14

[25] https://www.clarin.eu/blog/vlo-updated-advanced-search-facilities
[26] Piwki User Guide (http://piwik.org/docs/), Google Analytics (https://support.google.com/analytics#topic=3544906)

The result was similar, but new problems are found (Figure 26). This time Piwik successfully recorded the referrer website, but it failed to document the two visited web pages. There is also no way to record the activities of a web page (VLO record) when tabs are deployed inside (recorded as “3x”). That will catch an attention of developers when taking web design into account. Somehow downloads are also undetected, casting some doubts of recording reliability. In general conclusion, the quality of Piwik web tracking is acceptable, but there are some hints of unexpected outcomes or irregularities, implying a need of caveats when interpreting the data. On a more positive side, the experiments proved that the visitor log can be anonymously investigated further to understand what a particular user and/or user group look for, and what the potential needs are. In addition, Piwik offers User ID function[27], combining the registered users of the website and IP addresses to identify specific users and groups. If VLO implements user registration in the future to provide more user customised experience, it could be used to segment the VLO users more accurately. In any case, if Piwik is adequately used, serious marketing research can be carried out and business and market potential can be unlocked.

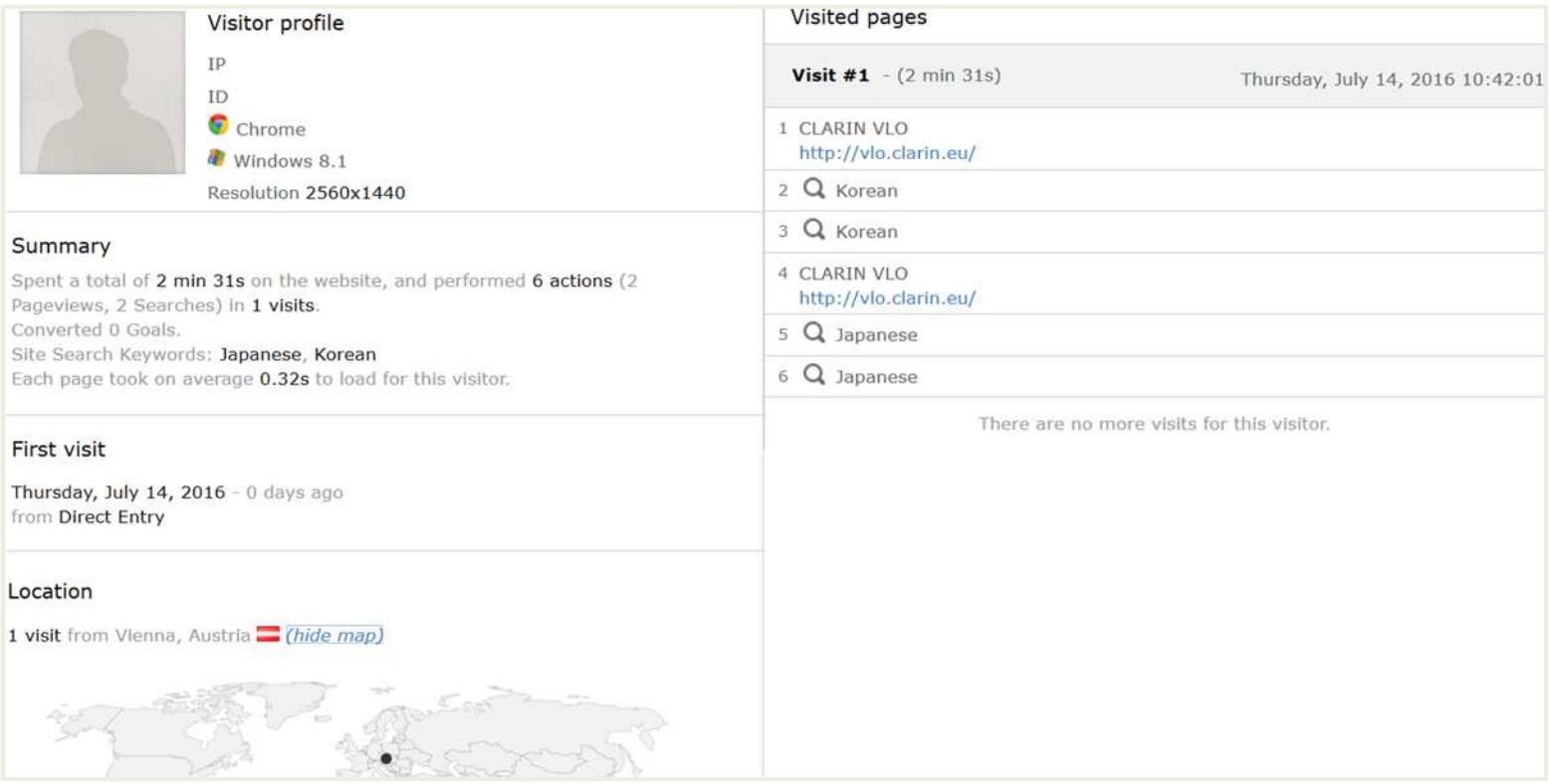

Figure 24. Visitor profile with the log on the right
(IPs are removed for privacy reasons)

[27] https://piwik.org/docs/user-id/

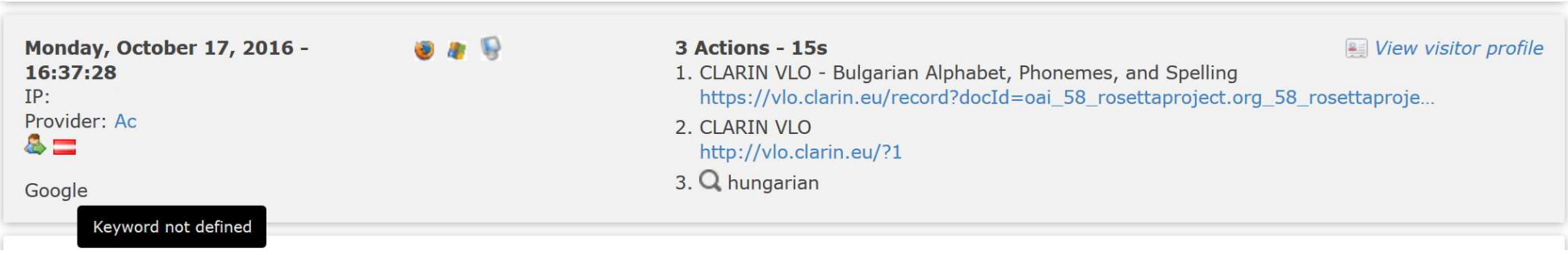

Figure 25. Visitor logs tracing the first user behaviours
(IPs are removed for privacy reasons)

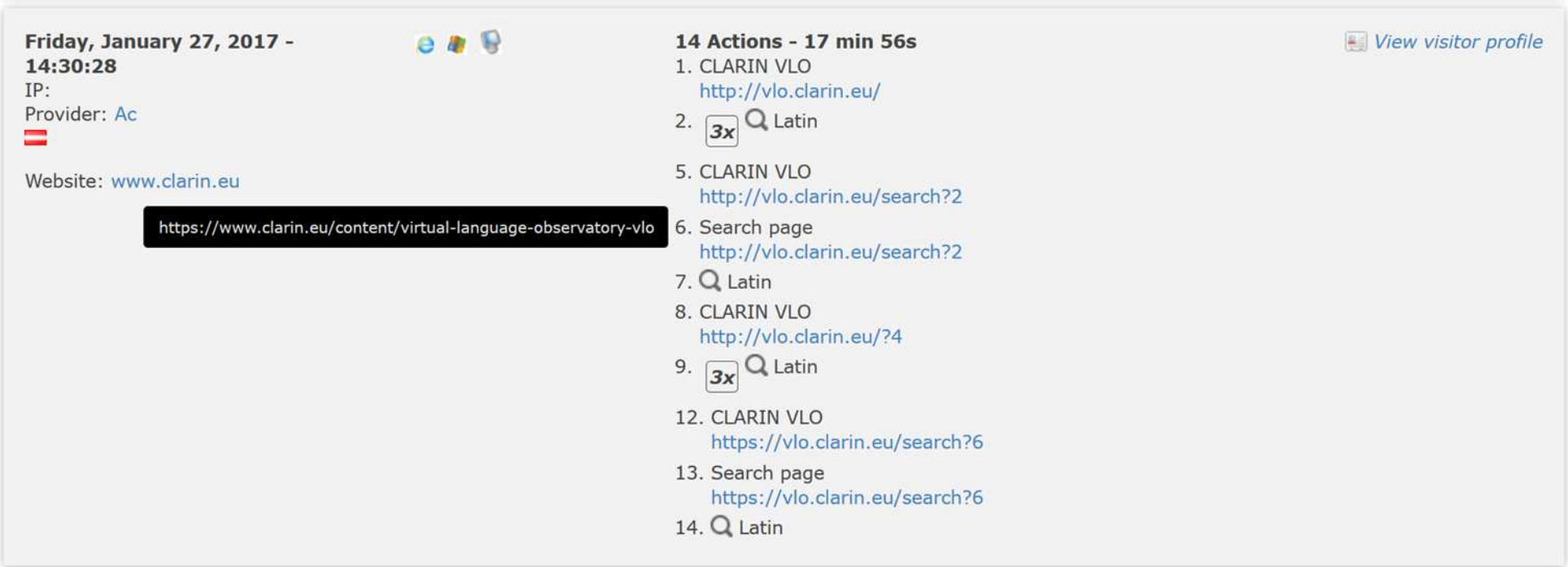

Figure 26. Visitor logs tracing the second user behaviours
(IPs are removed for privacy reasons)

## 4.2 Tips for data analysis

Although some may need a little training, Piwik is, in general, an intuitive and user-friendly tool. It can record and generate many different statistics about the use of websites. At the same time, it seems also true that many analysts only use the most basic (and default) functionalities without making effort to take various aspects of statistics into account, thus do not take advantage of its full potential. As a result, the data is sometimes neither adequately interpreted nor presented to the stakeholders. This would be also a consequence of the current research environment in which time may not be invested enough for marketing and business analyses. This section documents some tips for web analytics found during our analysis in section 3. The concept of *Open Evaluation* that this paper proposes is a little attempt to draw attention to this issue and break the barrier of access to the accumulated experience of web analytics. In addition, a broad category of Open Science and Innovation also fall within a scope of this idea to stimulate knowledge transfer and exchange. Although rather simple, the documentation of the use of Google Analytics by Szajewski (2013) is considered as a similar effort. He describes the step-by-step process of analysis on a university digital media repository. It is hoped that *Open Evaluation* would open an avenue to increase transparency, democracy, and competitiveness of such a research.

### 4.2.1 Segmentation

The default statistics are good way to know the overall trend, but they should not be interpreted without carefully examining smaller segments. As repeatedly stated in the section 3.4 and 3.5, user profiles and behaviours are diverse in different user groups. Thanks to the faceting function of Piwik, we can easily specify particular groups by a wide set of criteria and regular expressions, including IP address (Figure 27). As seen in section 4.1, we are even able to trace the behaviour of a unique user. For example, it is more than easy to identify the access from the CLARIN community in order to compare the CLARIN users with outsiders, as far as IP addresses are provided. Such a study is rare in the Digital Humanities and cultural heritage. Adding to the segmentation, the chronological analysis of segments has a lot of potential for the marketing analysis. The section 3 taught us the monitoring possibility of any segments, tracking the situation change over time (called “row evolution” in Piwik). It could open a door to validate hypotheses and drill down the users of the segments.

There are, however, some pitfalls too. The present setup of VLO perplexes Piwik. Sometimes it is not capable of clustering (called “flattening” in Piwik) VLO webpage segments as it fails to aggregate the URL variants of the same web pages. For example, due to the technical configurations, VLO uses a variety of session URLs including vlo/search, vlo/serarch?1, and vlo/search?9, which are all the same home page of the search engine. However, Piwik counts them separately on a random ba-sis. The consequence is the duplication of segments in the analytics. It makes it hard to obtain the entirely accurate statistics and can be only solved with a lot of manual “normalisation” effort.

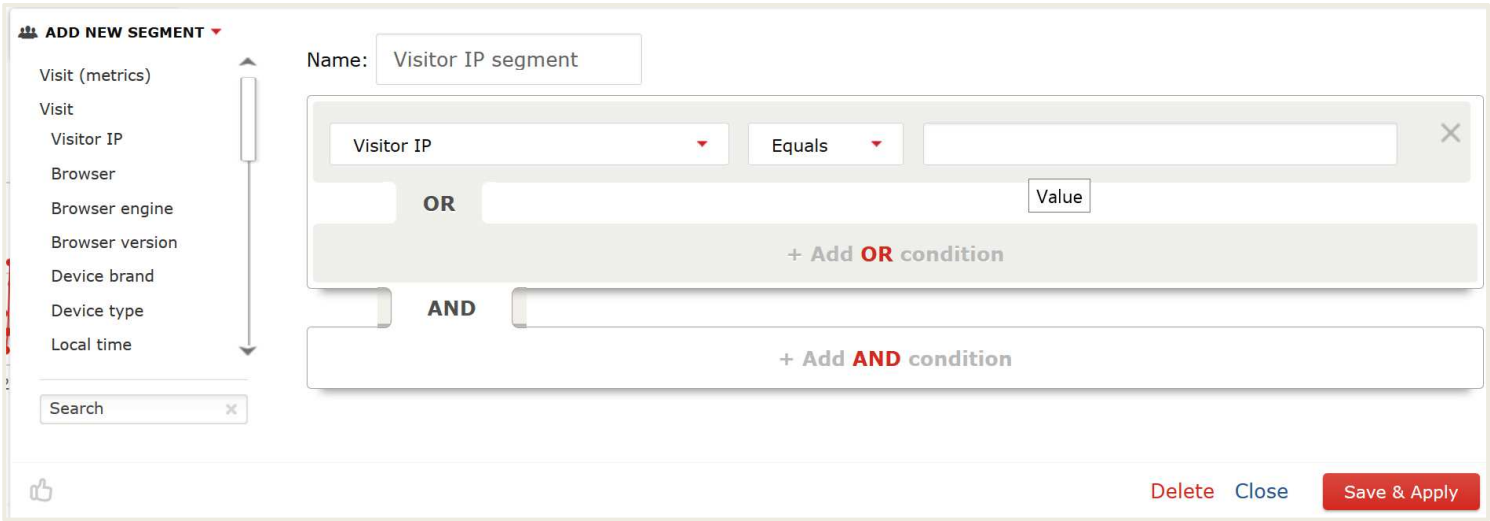

Figure 27. segmentation configuration in Piwik

### 4.2.2 Advanced Piwik and external applications

Default setting is more than enough for basic reporting, but the tool offers more complicated analysis instruments. One example is transition[28]. One can analyse the transition through a web page s/he selects (Figure 28). For instance, it is possible to examine how the users landed at a VLO record page (e.g. via internal link or search keyword “lemma”, or through an external search engine) and to where they go afterwards (e.g. back to the search result page, download, or exit). This paper can by no means address all the cases, but the detailed investigation is feasible, if required. Another case is a page overlay function[29]. It lets us check how links perform at the actual webpage layout (Figure29). Unfortunately it cannot be applied for the duration of our analysis period, therefore, we limit ourselves to just an introduction. Nonetheless, this function would contribute to a usability study of VLO.

As argued in the section 4.1, Piwik was unable to track the movements of a user within a webpage such as a click of a tab menu. This can be fixed with a Javascript event tracking code[30]. It can record, for example, that the actions of a user such as closing, maximizing, or minimising of a window. As such, Piwik is highly customisable. If such statistics are found important for usability and user experience, it is good to invest a little time to adapt the code.

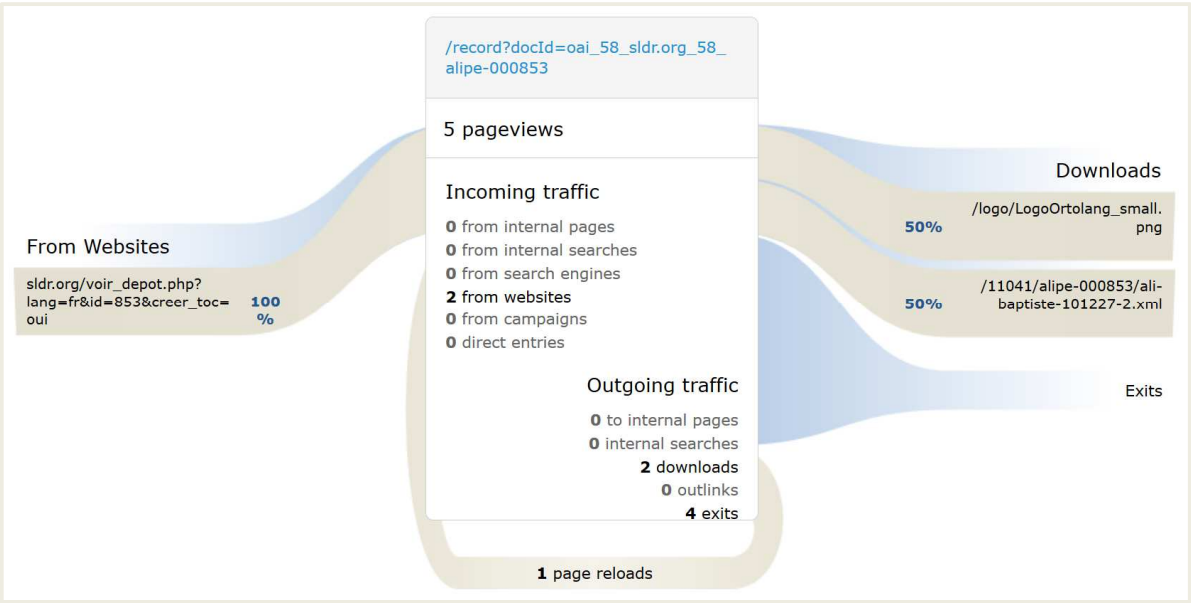

Figure 28. Transition report of a VLO record

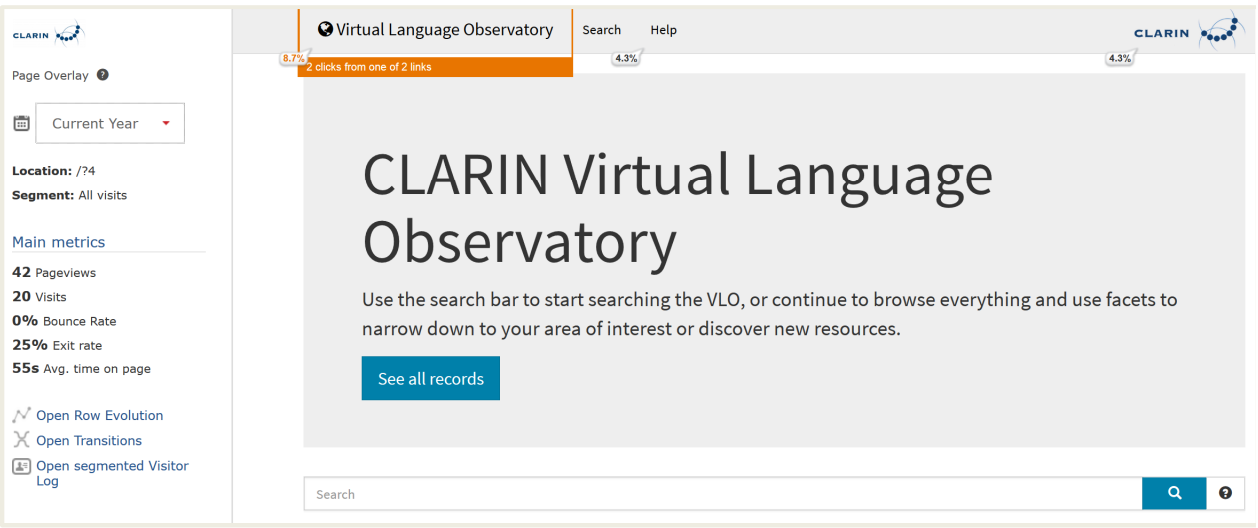

Figure 29. Page overlay showing the link statistics

We are not able to process, analyse, and visualise data with Piwik as flexible as other data analysis packages. In order to go beyond the default view and visualisation, raw data should be exported in other formats (CSV, TSV, XML, JSON, PHP) to manipulate with other tools. We already introduced a few tools in the previous sections, but we explain a little more how they are used.

Spreadsheet applications are more favoured to Piwik for data analysis and visualisation. As provided for Figure 23, the Pivot Table of EXCEL proves to be effective to discover previously unknown patterns. Another function which would be useful is correlation. In the past the author produced several analysis reports for APEx project, using basic correlation models to demonstrate the close relation between the archival contents and the access, which would be one of the success indicators of the project.

[28] https://piwik.org/docs/transitions/
[29] https://piwik.org/docs/page-overlay/
[30] http://piwik.org/docs/event-tracking/

Online tools such as the Geo Browser and Data Sheet Editor will help analysts to automate and visualise un-normalised data. The user simply has to copy and paste the place names found in Piwik into a web spreadsheet (Figure 30), which automatically detects them and assigns global coordinates. Then, they can be seamlessly plotted on a map in the Geo-Browser (Figure 9).

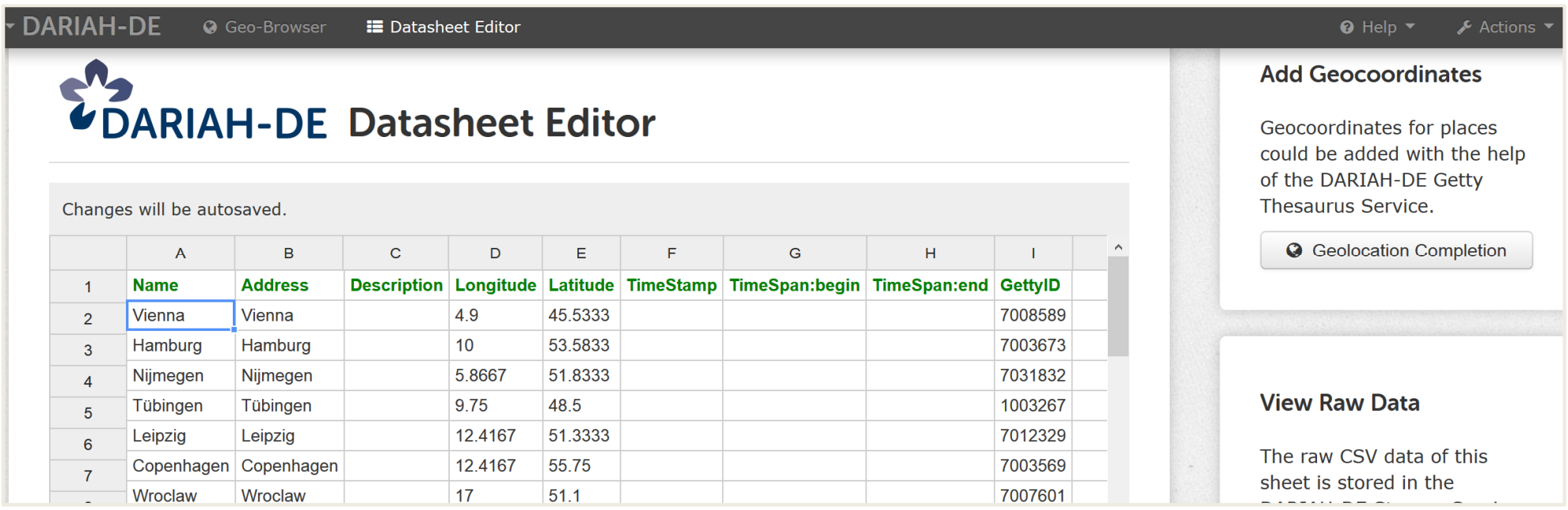

| | A | B | C | D | E | F | G | H | I |
|---|---|---|---|---|---|---|---|---|---|
| 1 | Name | Address | Description | Longitude | Latitude | TimeStamp | TimeSpan:begin | TimeSpan:end | GettyID |
| 2 | Vienna | Vienna | | 4.9 | 45.5333 | | | | 7008589 |
| 3 | Hamburg | Hamburg | | 10 | 53.5833 | | | | 7003673 |
| 4 | Nijmegen | Nijmegen | | 5.8667 | 51.8333 | | | | 7031832 |
| 5 | Tübingen | Tübingen | | 9.75 | 48.5 | | | | 1003267 |
| 6 | Leipzig | Leipzig | | 12.4167 | 51.3333 | | | | 7012329 |
| 7 | Copenhagen | Copenhagen | | 12.4167 | 55.75 | | | | 7003569 |
| | Wroclaw | Wroclaw | | 17 | 51.1 | | | | 7007601 |

Figure 30. Datasheet Editor[31] automatically identifies place names and assigns global coordinates

To generate an appealing tag clouds, the users can download a keyword frequency list as CSV file. Then, an EXCEL function generates iterations of words in columns (Figure 31). Some data normalisation would be needed for example, in order to remove spaces for a keyword with multiple words. Finally, one can copy and paste the result into a free online service such as the Word Cloud Generator[32] (Figure 17 and 18).

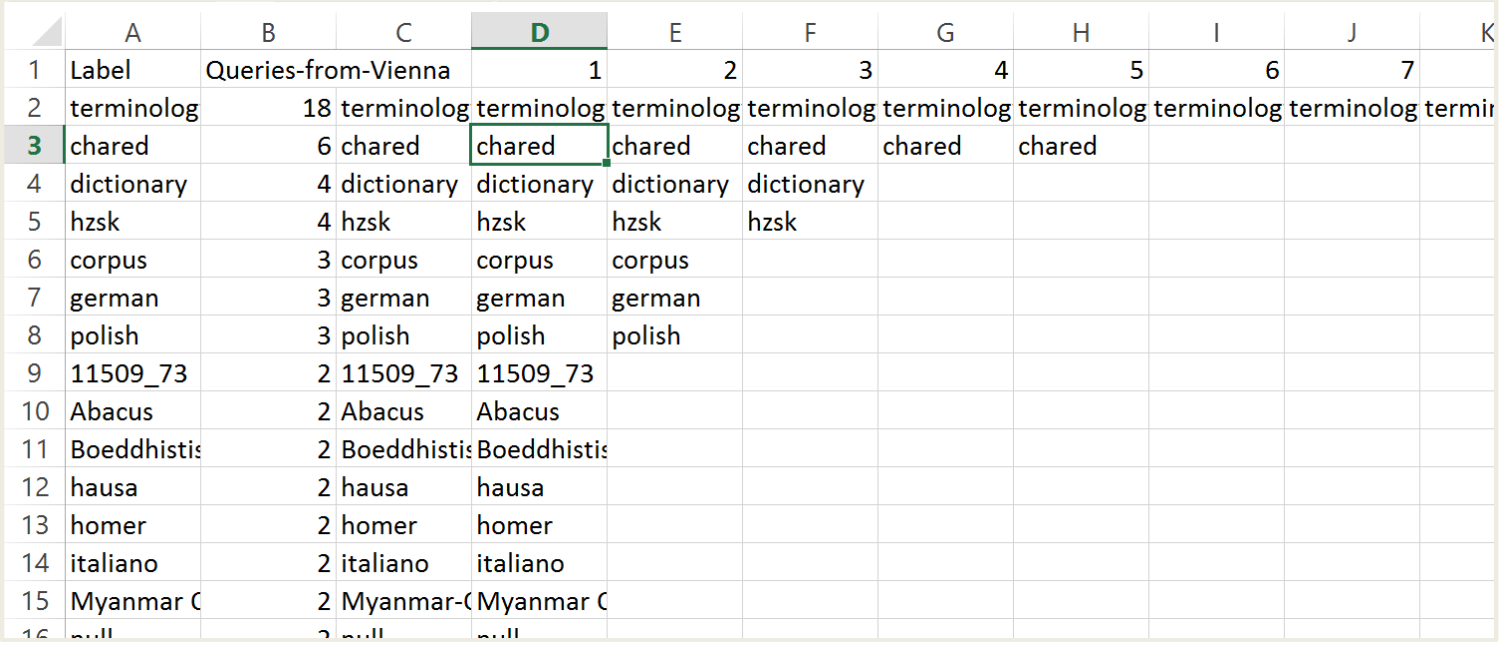

| | A | B | C | D | E | F | G | H | I | J | K |
|---|---|---|---|---|---|---|---|---|---|---|---|
| 1 | Label | Queries-from-Vienna | | 1 | 2 | 3 | 4 | 5 | 6 | 7 | |
| 2 | terminolog | 18 | terminolog | terminolog | terminolog | terminolog | terminolog | terminolog | terminolog | terminolog | termir |
| 3 | chared | 6 | chared | chared | chared | chared | chared | chared | | | |
| 4 | dictionary | 4 | dictionary | dictionary | dictionary | dictionary | | | | | |
| 5 | hzsk | 4 | hzsk | hzsk | hzsk | hzsk | | | | | |
| 6 | corpus | 3 | corpus | corpus | corpus | | | | | | |
| 7 | german | 3 | german | german | german | | | | | | |
| 8 | polish | 3 | polish | polish | polish | | | | | | |
| 9 | 11509_73 | 2 | 11509_73 | 11509_73 | | | | | | | |
| 10 | Abacus | 2 | Abacus | Abacus | | | | | | | |
| 11 | Boeddhistis | 2 | Boeddhistis | Boeddhistis | | | | | | | |
| 12 | hausa | 2 | hausa | hausa | | | | | | | |
| 13 | homer | 2 | homer | homer | | | | | | | |
| 14 | italiano | 2 | italiano | italiano | | | | | | | |
| 15 | Myanmar ( | 2 | Myanmar-( | Myanmar ( | | | | | | | |

Figure 31. Trick to manipulate CSV and convert it into appealing tag clouds

## 4.3 Myth or truth? –intuition, interpretation, and induction

In section 3, the author deliberately avoided delivering the last and only interpretation of data. Rather he subtly unpacks a series of steps taken, indicating the process of data analysis and interpretation. It is true that we have a common sense or intuitive interpretation when we first look at a dataset, because of previous experiences, therefore, we have a mutual understanding of "vague probability", in our case, about the web traffic. The problem is the proof of such interpretation is often missing and the process of induction is not documented. For example, we mentioned Wynne (2015b) in the introduction. To address this problem, one can find many traces of "analysis/research documentation". The section 3.1 explained why weekends are excluded from the graphs, while the section 3.2 examined the process and impact of Google indexing. The section 3.3 presented a potential pitfall of data visualisation and interpretation. A careful examination is needed in order not to be misguided by one interpretation or visualisation. The section 3.4 scratched the issues of observation and hypothesis, comparing two user groups: the whole group and the ÖAW user group. Despite a large volume of first-time and short visitors, relatively high average of visit duration and frequency seem to veil the heavy usage of some users. There is an instinctive hypothesis that the CLARIN partners would be the heavy users of VLO. Figure 13 to 16 proved it at least for the ÖAW. Figure 17 and 18 further discovered the different behaviours of the two user groups, which we cannot observe with the default statistics of the whole traf-fic. There is also a situation where comparison and interpretation is not sensible, due to the lack of data or the proportion of data (Figure 21 and 22). Such omission does not normally find a place in the final publication, but it might be valuable in the future research for whatever reason.

[31] https://geobrowser.de.dariah.eu/edit/
[32] https://www.jasondavies.com/wordcloud/

This paper is only able to work on intuitive hypotheses about our users, but ideally hypotheses should be also deployed in the VLO development. For instance, Fang (2007) analysed the statistics of Google Analytics and modified user interface based on three hypotheses. Comparing the web traffic of pre and post redesign of the law library website, he concluded that the new design brought more traffic, loyalty, and better navigation for the users. CLARIN can easily clone his simple methodology and use the objective evidence for not only data interpretation, but also actual improvement of our services.

It is equally important to acknowledge that this is the experiment of this paper attempting to contribute to the transparency of the research induction process. In this way *Open Evaluation* comes into play in order to illuminate a way of new research methodology and mind set in the digital century, because current research practices often fail to share the documentation of their processes.

# 5 Conclusion

## 5.1 Future Work

It would be very interesting to align the outcome of this paper with the statistics of the communication services of CLARIN. Although the author emphasised the synergy between marketing and software development, this alignment of such analyses was not conducted on this occasion. The author has, in fact, a substantial experience in his previous APEx project to produce in-depth quarterly reports for the web access of the Archives Portal Europe (the main user service of APEx project)[33] and the project website (communication website)[34] over two years. The internal-only reports showed fruitful comparison between the two websites, providing essential information about the strategic positions of different types of websites. They are created per each European country, as well as Europe and the world, and the participating project institutions are informed of the country reports. This report distribution strategy enables them to be aware of the situation of their country and to use the reports as an aid for their communication and outreach activities. Such knowledge transfer can be easily applied for CLARIN to establish promotional strategies for each member country.

Another potential work would be an alignment and/or integration with the Curation Module (Ostojic et al. 2016). It systematically and automatically collects a wide range of statistics about the quality (and the quantity) of CMD profiles, metadata records, and the collections. As the discoverability of records and the access of records should have some kind of correlation, the web traffic of particular VLO records would be reflected by the resource discoverability of CMD records. The impact of the Curation Module and the metadata quality can be investigated. It is particularly promising if it will continuously record the statistics over time, when the connection between Piwik and the module will become more interlinked in the time-dimension.

Due to the change of focus, the plan of this paper has changed. It originally aimed to compare statistics of various types of CLARIN web applications such as Weblicht and Bavarian Archive for Speech Signals (BAS). In future such investigation offers multi-faceted views on the users and serves as a basis of forming broader strategies for CLARIN marketing and software development. In addition, it was initially planned to include the analysis of identification services. For example, it would contribute to evaluate the situation of authorised access especially in respect to academia. Moreover, it could check the validity of our theory that user authentication hampers access. Furthermore, ambitious research would be conducted to analyse the transition reports (Figure 28) which can monitor the user traffic of websites within the multi-layered CLARIN web structures. For example, it is possible to examine the dynamic user flows between websites including the CLARIN main website, the VLO, and the data providers which deliver their metadata to the VLO (CLARIN centre repositories).

More evaluations of CLARIN services have also started recently. The annual CLARIN conference 2016 hosted two relevant sessions: user involvement meeting and a workshop on usability of web services[35]. The participants of the sessions openly discussed about the burning issues and confirmed the continuation of their works. The latter actually resulted in a formal deliverable in CLARIN-PLUS project (Vare et al. 2016), although the VLO part is quite limited among all other CLARIN services. On the same occasion, a paper was presented for the user evaluation of the VLO pertaining to translation studies (Lušicky and Wissik 2016), employing questionnaires for students. It was an encouraging sign from the CLARIN members to raise awareness of the user-centric services. Collating the result of web analytics of this paper and non-web-based surveys such as user interviews and questionnaires, or even (re-)doing the parallel evaluations would be a potential future work. The comparison of different types of surveys can answer more specific questions about the users, markets, and the performance of our marketing and infrastructural services.

[33] http://www.archivesportaleurope.net/
[34] http://www.apex-project.eu

### 5.2 Challenges for CLARIN's infinity and beyond –A little Kaizen

This article presented a minimum set of user analysis, attempting to address the issues of the user-oriented development method with respect to the web traffic and to capitalise on easily available statistics to increase the marketing value of the VLO. The usefulness and potential of Piwik were revealed. In addition, the author intentionally documented the process of data analysis and interpretation for the sake of research reusability and reproducibility. Although such documentation aspect of the web analytics made this paper rather verbose, it is hoped that other researchers would benefit from it in the future.

The current problem of CLARIN's technical and business development is the absence of a simple PDCA methodology and the documentation of visions and business strategies. As far as the author recognizes, the Check is preliminary carried out by internal testing and feedback, concentrating on bug fixing and self-contained improvement. Also it is not connected to marketing strategies, if any. The outreach and user involvement groups often implement their plans (Wynne 2015a), but do not deeply concern about the impact of their precious activities and/or have little means or experience to measure it. The first issue, in fact, partly originates from the second issue. There is actually a question mark for the clear documentation for the vision and strategies of CLARIN. At the moment, CLARIN does not seem to provide up-to-date information about their concrete visions and strategies as well as financial statements on its public website at least. There are several (out-dated) relevant reports and plans such as value prepositions, cost estimations, and financial plans[36][37]. However, without a concrete financial reports which should include a balance sheet, and without strategies document including KPIs and milestones, it is extremely hard to properly evaluate the CLARIN. In other words, this paper is able to analyse the VLO in general, but is unable to execute a proper evaluation and make concrete recommendations for the VLO, because the benchmarks and targets are not yet clearly defined. It is thus highly recommended to first resolve such problems. Then, the results of this paper would provide very useful statistics about the performance of one of the main technical services of CLARIN, so that RoI can be further discussed.

It is neither the conclusion of this paper to merely criticise the state of VLO performance, nor to overemphasize the needs of user analysis. Rather it is hoped that it has given enough incentive for the CLARIN community to try to invest more on marketing and strategy building alongside the technical development and outreach. As the CLARIN expands and plays a leading role in the Digital Humanities, there is no doubt that user centric approach will be a major element of its operation. We have also tried to examine and understand the mechanism and implication of a web analytics tool, which is necessary to interpret the phenomena of the users. It hopefully enables the readers to find universal lessons instead of project-specific results of the internet traffic. In addition, this paper raises awareness of *Open Evaluation* under transparent management, which includes the sharing of user statistics. As seen in other initiatives, it is a growing concern for the credibility of CLARIN as open research infrastructure, which would also lead to Open Science and Innovation. It is the intension of the author to take an approach of a constructive criticism. This document is, thus, served as a beginning of a little Kaizen[38] to successfully "sustain their sustainability" in many areas and in every level of CLARIN.

[35] https://www.clarin.eu/node/4382
[36] https://www.clarin.eu/content/reports
[37] https://www.clarin.eu/content/clarin-plus-deliverables
[38] https://www.kaizen.com/about-us/definition-of-kaizen.html

There is no doubt that CLARIN has proved its value and potential for scholarly community and beyond for the last decade, and this paper is a challenging attempt to support and bring it to the next level for the coming years.